\documentclass[10pt,twocolumn,letterpaper]{article}

\usepackage{cvpr}
\usepackage{times}
\usepackage{epsfig}
\usepackage{graphicx}
\usepackage{amsmath}
\usepackage{amssymb}
\usepackage{shortbold}
\usepackage{booktabs}
\usepackage{verbatim}
\usepackage{xcolor}
\usepackage{marginnote}
\usepackage{footmisc}

\usepackage{multibib} %
\newcites{sup}{Supplementary References}

\usepackage{pifont}
\newcommand{\cmark}{\ding{51}}%
\newcommand{\xmark}{\ding{55}}%

\definecolor{darkred}{rgb}{0.75,0,0}
\definecolor{darkgreen}{rgb}{0,0.75,0}

\IfFileExists{darkmode.txt}{\usepackage{pagecolor} \pagecolor{darkgray} \color{lightgray}}{} %

\usepackage[pagebackref=true,breaklinks=true,letterpaper=true,colorlinks,bookmarks=false]{hyperref}
\hypersetup{
    plainpages=false,
    colorlinks=true,              %
    linkcolor=blue,               %
    anchorcolor=blue,             %
    citecolor=blue,               %
    filecolor=blue,               %
    urlcolor=blue,                %
    pdfview=FitH,                 %
    pdfstartview=FitH,            %
    pdfpagelayout=SinglePage      %
}

\cvprfinalcopy

\def\cvprPaperID{1360}

\ifcvprfinal\fi

\makeatletter
\DeclareRobustCommand\onedot{\futurelet\@let@token\@onedot}
\def\@onedot{\ifx\@let@token.\else.\null\fi\xspace}
\def\eg{\emph{e.g}\onedot}

\makeatother

\DeclareMathOperator*{\argmin}{arg\,min}
\DeclareMathOperator*{\cossim}{sim}

\definecolor{dark-green}{rgb}{0,0.4,0}

\begin{document}

\title{\textbf{Cross-task weakly supervised learning from instructional videos}}
\author{Dimitri Zhukov\textsuperscript{1,2}
\and Jean-Baptiste Alayrac\textsuperscript{1,3}
\and Ramazan Gokberk Cinbis\textsuperscript{4}
\and David Fouhey\textsuperscript{5}
\and Ivan Laptev\textsuperscript{1,2}
\and Josef Sivic\textsuperscript{1,2,6}
}

\date{\vspace{-5ex}}

\maketitle

\footnotetext[1]{Inria, France}
\footnotetext[2]{D\'{e}partement d'informatique de l'Ecole Normale Sup\'{e}rieure, PSL Research University, Paris, France}
\footnotetext[3]{Now at DeepMind}
\footnotetext[4]{Middle East Technical University, Ankara, Turkey}
\footnotetext[5]{University of Michigan, Ann Arbor, MI}
\footnotetext[6]{CIIRC -- Czech Institute of Informatics, Robotics and
Cybernetics at the Czech Technical University in Prague}

\begin{abstract}
In this paper we investigate learning visual models for the steps of ordinary tasks using weak supervision via instructional narrations and an ordered list of steps instead of strong supervision via temporal annotations. At the heart of our approach is the observation that weakly supervised learning may be easier if a model shares components while learning different steps: ``pour egg'' should be trained jointly with other tasks involving ``pour'' and ``egg''.  We formalize this in a component model for recognizing steps and a weakly supervised learning framework that can learn this model under temporal constraints from narration and the list of steps. Past data does not permit systematic studying of sharing and so we also gather a new dataset, CrossTask, aimed at assessing cross-task sharing. Our experiments demonstrate that sharing across tasks improves performance, especially when done at the component level and that our component model can parse previously unseen tasks by virtue of its compositionality.
\end{abstract}

\section{Introduction}
\label{sec:introduction}
Suppose you buy a fancy new coffee machine and you would like to make a latte.
How might you do this? After skimming the instructions, you may start
watching instructional videos on YouTube to figure out what each step entails:
how to press the coffee, steam the milk, and so on. In the process, you would
obtain a good visual model of what each step, and thus the entire task, looks like.
Moreover, you could use parts of this visual model of making lattes to help
understand videos of a new task, e.g., making filter coffee, since various nouns and verbs
are shared. The goal of this paper is to build automated systems that can similarly
learn visual models from instructional videos and in particular, make use of shared
information across tasks (e.g., making lattes and making filter coffee).

The conventional approach for building visual models of how to do things \cite{carreira17quovadis,simonyan2014two,Wang13action}
is to first annotate each step of each task in time and then train a
supervised classifier for each. Obtaining strong supervision in the
form of temporal step annotations is
time-consuming, unscalable and, as demonstrated by humans' ability to learn
from demonstrations, unnecessary. Ideally, the method should be
weakly supervised (i.e., like~\cite{Alayrac15Unsupervised, feifei2016connectionist, kuehne17weakly, Sener15unsupervised}) and jointly learn {\it when} steps
occur and {\it what} they look like. Unfortunately, any weakly supervised
approach faces two large challenges. Temporally localizing steps in the
input videos for each task is hard as there is a combinatorial set of
options for the step locations; and, even if the steps were localized,
each  visual model learns from limited data and may work poorly.

\begin{figure}
\includegraphics[width=\linewidth]{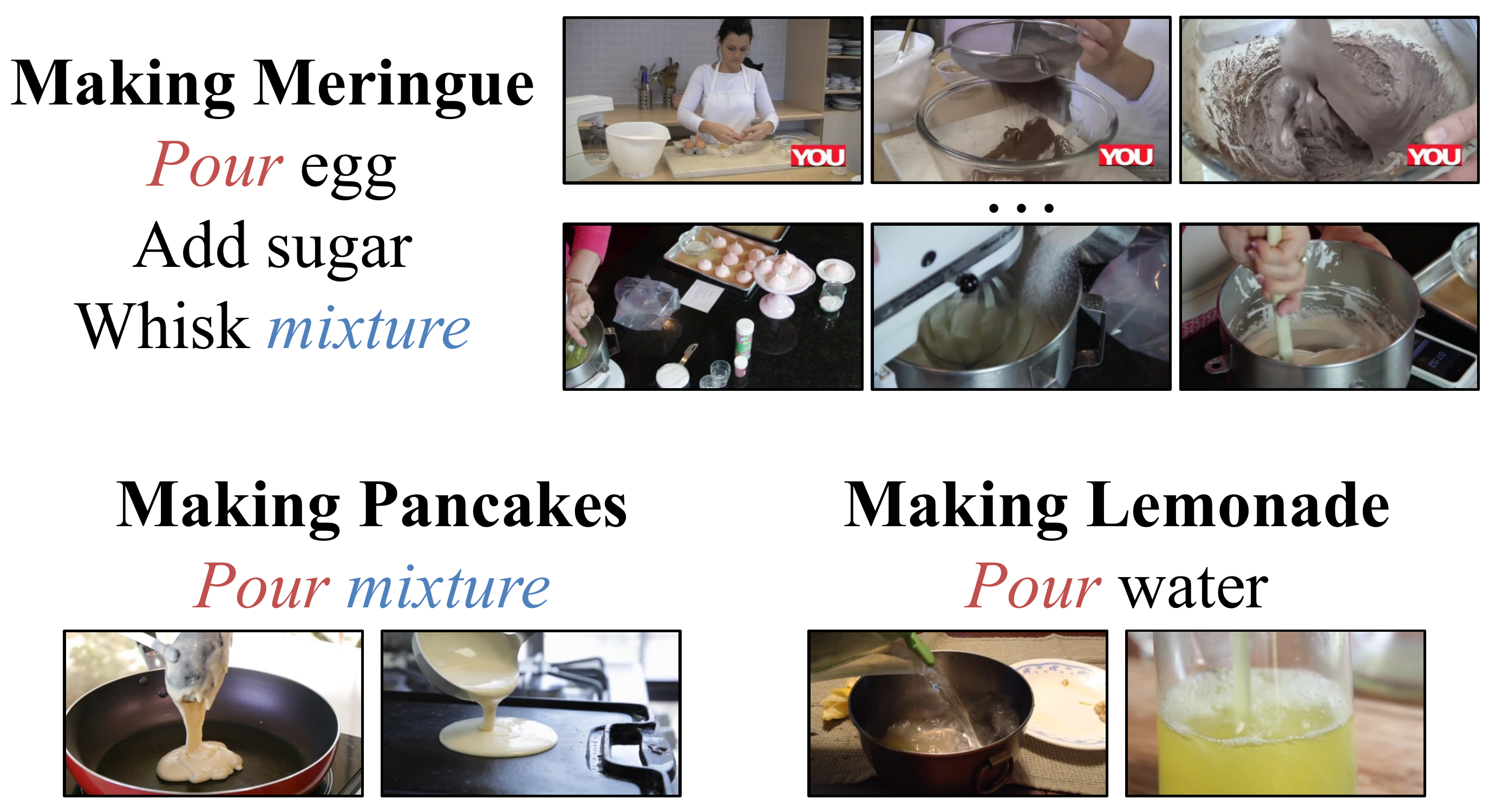}
\caption{Our method begins with a collection of tasks, each consisting of
an ordered list of steps and a set of instructional videos from YouTube.
It automatically discovers both where the steps occur
and what they look like. To do this, it uses the order, narration and commonalities
in appearance across tasks (e.g., the appearance of {\it pour}
in both {\it making pancakes} and {\it making meringue}).}
\label{fig:teaser}
\vspace{-0.1in}
\end{figure}

We show how to overcome these challenges by sharing across tasks and using
weaker and naturally occurring forms of supervision.  The related tasks let us
learn better visual models by exploiting commonality across steps as
illustrated in Figure~\ref{fig:teaser}.  For example, while learning about {\it
pour water} in {\it making latte}, the model for {\it pour} also depends on
{\it pour milk} in {\it making pancakes} and the model for {\it water} also
depends on {\it put vegetables in water} in {\it making bread and butter pickles}.
We assume an ordered list of steps is given per task and that the videos are
instructional (i.e., have a natural language narration describing what is being
done).  As it is often the case in weakly supervised video learning
\cite{alayrac16objectstates,feifei2016connectionist,Sener15unsupervised}, these assumptions
constrain the search for when steps occur, helping tackle a combinatorial search space.

We formalize these intuitions in a framework, described in Section \ref{sec:model}, that
enables compositional sharing across tasks together with temporal constraints for
weakly supervised learning. Rather than learning each step as
a monolithic weakly-supervised classifier, our formulation learns a component model
that represents the model for each step as the combination of models of its components, or
the words in each step (e.g., {\it pour} in {\it pour water}).
This empirically improves learning performance and these component
models can be recombined in new ways to parse videos for tasks for which
it was not trained, simply by virtue of their representation.
This component model, however, prevents the direct application of techniques
previously used for weakly supervised learning in similar settings
(e.g., DIFFRAC \cite{Bach07diffrac} in
\cite{alayrac16objectstates}); we therefore introduce a new and more general
formulation that can handle more arbitrary objectives.

Existing instructional video datasets do not
permit the systematic study of this sharing. We gather a new dataset, CrossTask,
which we introduce in Section \ref{sec:dataset}. This dataset consists
of 4.7K instructional videos for 83 different tasks, covering 376 hours of
footage. We use this dataset to compare our proposed approach with a number of
alternatives in experiments described in Section \ref{sec:experiments}.
Our experiments aim to assess the following three questions: how well does the
system learn in a standard weakly supervised setup; can it exploit related
tasks to improve performance; and how well can it parse previously unseen
tasks.

The paper's contributions include:
{\bf (1)} A component model that shares information between steps for
weakly supervised learning from instructional videos; {\bf (2)} A weakly
supervised learning framework that can handle such a model together with
constraints incorporating different forms of weak supervision; and {\bf (3)}
A new dataset that is larger and more diverse than past efforts, which
we use to empirically validate the first two contributions. We make our dataset and our code publically available\footnote{\url{https://github.com/DmZhukov/CrossTask}\label{fn:dataset}}.

\section{Related Work}
\label{sec:related_work}
Learning the visual appearance of steps of a task from
instructional videos is a form of action recognition.
Most work in this area, e.g.,
\cite{carreira17quovadis,simonyan2014two,Wang13action}, uses strong supervision
in the form of direct labels, including a lot of work that focuses on similar
objectives \cite{Damen2018,Fang2018,Fouhey18}.
We build our feature representations on top of advances in this area \cite{carreira17quovadis}, but our proposed method does not depend on having lots of annotated data for our problem.

\begin{figure*}[t]
\begin{center}
   \includegraphics[width=\linewidth]{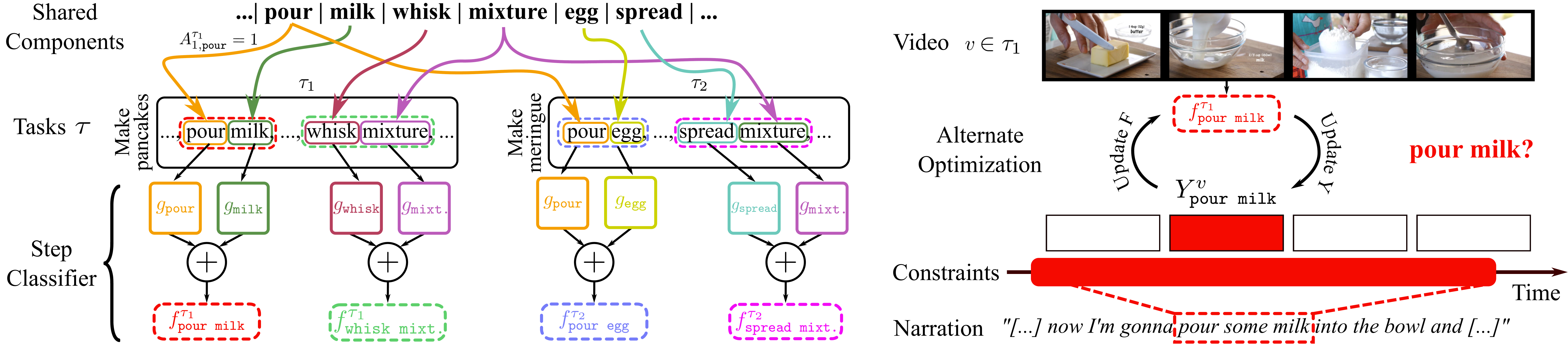}
\caption{Our approach expresses classifiers for each step of each task in terms
of a component model (e.g., writing the {\it pour milk} as a {\it pour} and
{\it milk} classifier). We thus cast the problem of learning
the steps as learning an underlying set of component models. We learn these
models by alternating between updating labels for these classifiers and the
classifiers themselves while using constraints from narrations. }
\label{fig:method}
\vspace{-0.25in}
\end{center}
\end{figure*}

We are not the first to try to learn with weak supervision in videos and our
work bears resemblances to past efforts. For instance, we make use of ordering
constraints to obtain supervision, as was done in~\cite{Bojanowski14weakly,feifei2016connectionist,kuehne17weakly,richard17weakly,Bojanowski15weakly}.
The aim of our work is perhaps closest to~\cite{Alayrac15Unsupervised,Malmaud15what,Sener15unsupervised} as they also use narrations in the context of instructional videos.
Among a number of distinctions with each individual work, one significant
novelty of our work is the compositional model used, where instead of learning
a monolithic model independently per-step as done
in~\cite{Alayrac15Unsupervised, Sener15unsupervised}, the framework shares
components (e.g., nouns and verbs) across steps. This sharing improves performance, as we empirically confirm, and enables the parsing of unseen tasks.

In order to properly evaluate the importance of sharing, we gather a dataset of
instructional videos. These have attracted a great deal of attention recently
\cite{Alayrac15Unsupervised,alayrac16objectstates,huang17unsupervised,huang18finding,Malmaud15what,Sener15unsupervised,zhou18towards}
since the co-occurrence of demonstrative visual actions and natural language
enables many interesting tasks ranging from coreference
resolution~\cite{huang17unsupervised} to learning person-object
interaction~\cite{alayrac16objectstates,dima2014youdo}. Existing data, however, is either
not large (e.g., only 5 tasks \cite{alayrac16objectstates}), not diverse (e.g.,
YouCookII~\cite{zhou18towards} is only cooking), or not densely temporally annotated
(e.g., What's Cooking?~\cite{Malmaud15what}).
We thus collect a dataset that is: {\bf (i)} relatively large (83 tasks, ~4.7K videos);
{\bf (ii)} simultaneously diverse (Covering car maintenance, cooking, crafting)
yet also permitting the evaluation of sharing as it has
related tasks; and {\bf (iii)} annotated for temporal localization,
permitting evaluation.  The scale, and relatedness, as we demonstrate
empirically contribute to increased performance of visual models.

Our technical approach to the problem builds particularly heavily on the use of
discriminative clustering~\cite{Bach07diffrac,Xu2004maximum}, or the simultaneous constrained
grouping of data samples and learning of classifiers for groups.
Past work in this area has either had operated with complex constraints and a restricted classifier (e.g., minimizing the L2 loss with linear model
\cite{Bach07diffrac,alayrac16objectstates}) or an unrestricted classifier, such as a deep network, but no constraints
\cite{bojanowski17unsupervised,Caron2018}. Our weakly supervised setting requires the ability to add constraints
in order to converge to a good solution while our compositional model and desired loss function requires the ability to use an unrestricted classifier.
We therefore propose an optimization approach that handles both, letting us train with a compositional model while also using
temporal constraints.

Finally, our sharing between tasks is enabled via the composition of the components of each step (e.g., nouns, verbs).
This is similar to attributes \cite{Farhadi09,Ferrari07}, which have been used in action recognition in the past \cite{Liu11,Yao11}.
Our components are meaningful (representing, e.g., ``lemon'') but also
automatically built; they are thus different than pre-defined semantic attributes (not automatic)
and the non-semantic attributes (not intrinsically meaningful) as defined in \cite{Farhadi09}.
It is also related to methods that compose new classifiers from others, including
\cite{misra2017composing,Yatskar_Commonly_17,guadarrama13} among many others. Our framework is orthogonal, and shows how to learn these in a weakly-supervised setting.

\section{Overview}
\label{sec:overview}
Our goal is to build visual models for a set of {\bf tasks} from
instructional videos.  Each task is a multi-step process such as
{\it making latte} consisting of multiple {\bf steps}, such as {\it pour milk}.
We aim to learn a visual model for each of these steps. Our approach
uses {\bf component models} that represent each step in terms of
its constituent {\bf components} as opposed to a monolithic entity, as
illustrated in Figure~\ref{fig:method}. For instance, rather than building a classifier
solely for {\it whisk mixture} in the context of {\it make pancakes}, we learn
a set of classifiers per-component, one for {\it whisk}, {\it spread}, {\it mixture} and so on, and
represent {\it whisk mixture} as the combination of {\it whisk} and {\it mixture}
and share {\it mixture} with {\it spread mixture}.
This shares data between steps and enables the parsing of previously unseen
tasks, which we both verify empirically.

We make a number of assumptions.
Throughout, we assume that we are given an ordered list of steps for each task.
This list is our only source of manual supervision and is done once per-task and
is far less time consuming than annotating a temporal segmentation of each step
in the input videos. At training time,
we also assume that our training videos contain audio that explains what actions are being performed. At test time, however,
we do not use the audio track: just like a person who watches a video online, once
our system is shown how to make a latte with narration, it is expected to follow along
without step-by-step narrations.

\section{Modeling Instructional Videos}
\label{sec:model}
We now describe our technical approach for using a list of steps
to jointly learn the labels and visual models on a set of
narrated instructional videos. This is weakly supervised since
we provide only the list of steps, but not their temporal locations
in training videos.

\noindent {\bf Problem formulation.}
We denote the set of narrated instructional videos $\mathcal{V}$. Each video
$v\in\mathcal{V}$ contains a sequence of $N_v$ segments of visual features $X^v=(x_1, \ldots, x_{N_v})$
as well as narrations we use later.
For every task $\tau$ we assume to be given a set of videos $V_\tau$
together with a set of ordered natural language steps $K_\tau$.

Our goal is then to discover a set of classifiers $F$ that can identify the steps of the tasks.
In other words, if $\tau$ is a task and $k$ is its step, the classifier $f_k^\tau$
determines whether a visual feature depicts step $k$ of $\tau$ or not.
To do this, we also learn a labeling $Y$ of the
training set for the classifiers, or for every video $v$ depicting task $\tau$,
a binary label matrix $Y^v\in\{0,1\}^{N_v \times K_\tau}$ where $Y^v_{tk} = 1$ if
time $t$ depicts step $k$ and $0$ otherwise. While jointly learning labels and
classifiers leads to trivial solutions, we can eliminate these and make
meaningful progress by constraining $Y$ and by sharing information across the
classifiers of $F$.

\subsection{Component Classifiers}

One of the main focuses of this paper is in the form of the step classifier $f$.
Specifically, we propose a component model that represents each step (e.g., ``pour milk'')
as a combination of components (e.g., ``pour'' and ``milk''). Before explaining how
we formulate this, we place it in context by introducing a variety of alternatives that
vary in terms of how they are learned and formulated.

The simplest approach, a {\bf task-specific step model}, is to learn a
classifier for each step in the training set (i.e., a model for {\it pour egg} for the
particular task of {\it making pancakes}). Here, the model simply learns
$\sum_\tau K_\tau$ classifiers, one for each of the $K_\tau$ steps in each
task, which is simple but which permits no sharing.

One way of adding sharing would be to have a {\bf shared step model}, where a single
classifier is learned for each unique step in the dataset. For instance, the
{\it pour egg} classifier learns from both {\it making meringues} and {\it making
  pancakes}.
This sharing, however, would be limited to exact duplicates
of steps, and so while {\it whisk milk} and {\it pour milk} both share an object,
they would be learned separately.

Our proposed {\bf component model} fixes this issue. We automatically generate a vocabulary
of {\bf components} by taking the set of stemmed words in all the
steps. These components are typically objects, verbs and prepositions and we combine classifiers for each component to yield our steps.
In particular, for a vocabulary of $M$ components, we
define a per-task matrix $A^\tau \in \{0,1\}^{K_\tau \times M}$ where
$A^\tau_{k,m} = 1$ if the step $k$ involves components $m$ and $0$ otherwise.
We then learn $M$ classifiers $g_1, \ldots, g_M$ such that
the prediction of a step $f^\tau_k$ is the average of predictions provided by component classifiers
\begin{equation}
\label{eq:attr_clf}
f^\tau_k(x)=\sum\limits_{m}A^\tau_{km}g_m(x) / \sum\limits_{m}A^\tau_{km}.
\end{equation}
For instance,
the score for {\it pour milk} is the average of outputs of $g_{\textit{pour}}$
and $g_{\textit{milk}}$. In other words, when optimizing over the set
of functions $F$, we optimize over the parameters of $\{g_i\}$ so that
when combined together in step models via (\ref{eq:attr_clf}), they produce
the desired results.

\subsection{Objective and Constraints}
\label{subsec:constraints}

Having described the setup and classifiers, we now describe the objective
function we minimize. Our goal is to simultaneously optimize over step location
labels $Y$ and classifiers $F$ over all videos and tasks
\begin{equation}
\label{eq:main}
\min_{Y\in \mathcal{C}, F\in\mathcal{F}}\sum_{\tau}\sum_{v \in \mathcal{V}(\tau)} h(X^v,Y^v; F),
\end{equation}
where $\mathcal{C}$ is the set of temporal constraints on $Y$ defined below, and
$\mathcal{F}$ is a family of considered classifiers. Our objective
function per-video is a standard cross-entropy loss
\begin{equation}
\label{eq:loss_function}
h(X^v,Y^v; F)=-\sum\limits_{t,k}Y^v_{tk}\log\left(\frac{\exp\left(f^\tau_k(x^v_t)\right)}{\sum\limits_{k'}\exp(f^\tau_{k'}(x^v_t))}\right).
\end{equation}

Optimizing (\ref{eq:main}) may lead to trivial solutions (e.g., $Y^v = 0$ and
$F$ outputting all zeros). We thus constrain our labeling of $Y$
to avoid this and ensure a sensible solution. In particular, we
impose three constraints:

\noindent {\bf At least once.} We assume that every video $v$ of a task depicts each step $k$
at least once, or $\sum_{t} Y_{tk}^v \ge 1$. \\
\noindent {\bf Temporal ordering.} We assume that steps occur in the given order. While not always
strictly correct, this dramatically reduces the search space and leads to better classifiers.\\
\noindent {\bf Temporal text localization.} We assume that the steps and corresponding narrations happen close in time,
e.g., the narrator of a {\it grill steak} video may say
``just put the marinated steak on the grill''.
We automatically compare the text description of each step
to automatic YouTube subtitles. For a task with $K_\tau$ steps and a video with $N_v$ frames,
we construct a $[0,1]^{N_v \times K_\tau}$ matrix of cosine similarities between steps and
a sliding-window word vector representations of narrations (more details in supplementary materials).
Since narrated videos contain spurious mentions of tasks (e.g., "before putting the steak on
the grill, we clean the grill") we do not directly use this matrix, but instead
find an assignment of steps to locations that maximizes the total similarity while respecting the
ordering constraints. The visual model must then more precisely identify when the action appears.
We then impose a simple hard constraint of disallowing labelings $Y^v$ where any step
is outside of the text-based interval (average length 9s)

\subsection{Optimization and Inference}

We solve problem~\eqref{eq:main} by alternating between updating assignments $Y$ and the parameters of the classifiers $F$.
\\
\noindent\textbf{Updating $Y$.}
When $F$ is fixed, we can minimize \eqref{eq:main} w.r.t. $Y$ independently for each video.
In particular, fixing $F$ fixes the classifier scores, meaning
that minimizing \eqref{eq:main} with respect to $Y^v$ is a constrained minimization of a linear
cost in $Y$ subject to constraints. Our supplemental shows that this can be done by dynamic programming.

\noindent\textbf{Updating $F$.}
When $Y$ is fixed, our cost function reduces to a standard supervised
classification problem. We can thus apply standard techniques for solving
these, such as stochastic gradient descent. More details are provided below and in the supplemental material.
\\
\noindent\textbf{Initialization.} Our objective is non-convex and has local minima,
thus a proper initialization is important.  We obtain such
an initialization by treating all assignments that satisfy the temporal text
localization constraints as ground-truth and optimizing for $F$ for 30
epochs, each time drawing a random sample that satisfies the constraints.

\noindent {\bf Inference.} Once the model has been fit to the data, inference on a new
video $v$ of a task $\tau$ is simple. After extracting features, we run each
classifier $f$ on every temporal segment, resulting in a $N_v \times K_\tau$ score matrix.
To obtain a hard labeling, we use dynamic programming to find the
best-scoring labeling that respects the given order of steps.

\begin{figure*}[t]
\includegraphics[width=\linewidth]{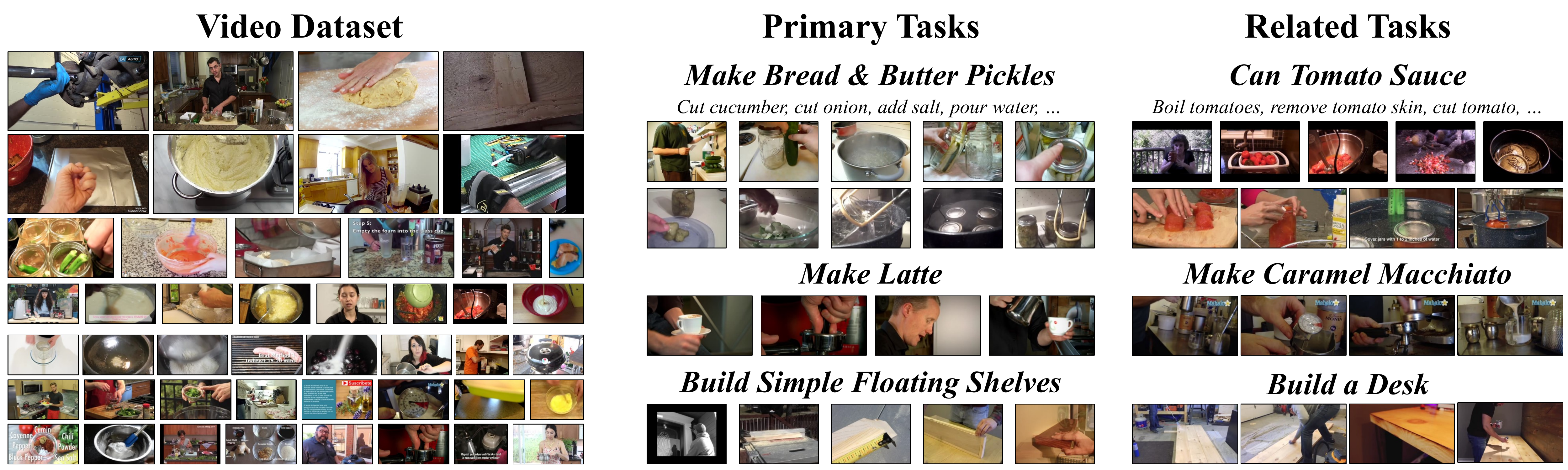}
\caption{Our new dataset, used to study sharing in a weakly supervised learning setting.
It contains primary tasks, such as {\it make bread and butter pickles}, as well as
related tasks, such as {\it can tomato sauce}. This lets us study whether learning
multiple tasks improves performance.}
\vspace{-0.1in}
\end{figure*}

\subsection{Implementation Details}

{\it Networks:} Due to the limited data size and noisy supervision, we use a linear classifier with dropout for regularization.
Preliminary experiments with deeper models did not yield improvements.
We use ADAM \cite{adam} with the learning rate of $10^{-5}$ for optimization.
{\it Features:} We represent each video segment $x_i$ using RGB I3D features
\cite{carreira17quovadis} (1024D), Resnet-152 features \cite{he16resnet} (2048D) extracted at
each frame and averaged over one-second temporal windows, and audio features from
\cite{hershey17audio} (128D).
{\it Components:} We obtain the dictionary of components by finding the set of unique stemmed words over all step descriptions. The total number of components is 383.
{\it Hyperparameters:} Dropout and the learning rate are chosen on a validation data set.

\section{CrossTask dataset}
\label{sec:dataset}
\begin{table}
\centering
\caption{A comparison of CrossTask with existing instructional datasets. Our dataset is both large and more diverse while also having temporal annotations.}
\label{tab:datatable}
{\small
\setlength\tabcolsep{4.7pt}
\begin{tabular}{lccccc} \toprule
~ & Num. & Total & Num. & Not only & Avail. \\
~ & Vids & Length & Tasks & Cooking & Annots \\ \midrule
\cite{alayrac16objectstates} & 150 & 7h & 5 & \textcolor{darkgreen}{\cmark} & Windows
\\
\cite{Sener15unsupervised} & 1.2K+85 & 100h & 17 & \textcolor{darkgreen}{\cmark} & Windows
\\
\cite{zhou18towards} & 2K & 176h & 89 & \textcolor{darkred}{\xmark} & Windows
\\
\cite{Malmaud15what} & 180K & 3,000h & \textcolor{darkred}{\xmark} & \textcolor{darkred}{\xmark} &
Recipes
\\
CrossTask & 4.7K & 376h & 83 & \textcolor{darkgreen}{\cmark} & Windows
\\
\bottomrule
\end{tabular}
}
\vspace{-0.1in}
\end{table}

One goal of this paper is to investigate whether sharing improves
the performance of weakly supervised learning from instructional videos.
To do this, we need a dataset covering a diverse set of interrelated tasks and annotated with
temporal segments. Existing data fails to satisfy at least one of these
criteria and we therefore collect a new dataset (83 tasks, 4.7K videos) related
to cooking, car maintenance, crafting, and home repairs.
These tasks and their steps are derived
from wikiHow, a website that describes how to solve many tasks,
and the videos come from YouTube.

CrossTask dataset is divided into two sets of tasks to investigate sharing. The first is {\bf primary tasks},
which are the main focus of our investigation and the backbone of the dataset.
These are fully annotated and form the basis for our evaluations.
The second is {\bf related tasks} with videos gathered in a more
automatic way to share some, but not all, components with the primary tasks.
One goal of our experiments is to assess whether these related
tasks improve the learning of primary tasks, and whether one can learn a
good model only on related tasks.

\subsection{Video Collection Procedure}
\label{subsec:video_collection}

\begin{figure*}[t]
\includegraphics[width=\linewidth]{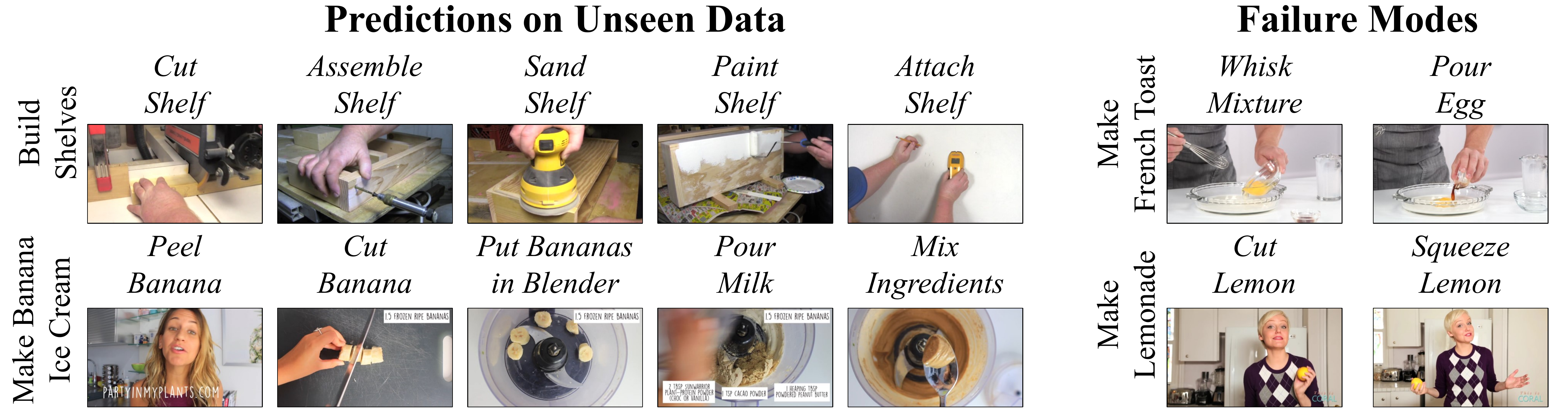}
\caption{Predictions on unseen data as well as typical failure modes. Our method does well on steps with distinctive motions and appearances.
Failure modes include (top) features that cannot make
fine-grained distinctions between e.g., egg and vanilla extract; and
(bottom) models that overreact to particular nouns, preferring a more visible lemon over
a less visible lemon actually being squeezed.}
\label{fig:exp_qualitative}
\vspace{-0.1in}
\end{figure*}

We begin the collection process by defining our tasks. These must satisfy
three criteria: they must entail a sequence of physical interactions
with objects (unlike e.g., {\it how to get into a relationship}); their
step order must be deterministic (unlike
e.g., {\it how to play chess}); and they must appear frequently
on YouTube. We asked annotators to review the tasks in five sections of wikihow
to get tasks satisfying the first two criteria, yielding $\sim 7$K candidate tasks,
and manually filter for the third criteria.

We select 18 primary tasks and 65 related tasks from these $7$K candidate tasks. The
primary tasks cover a variety of themes (e.g.,  auto repair to
cooking to DIY) and include
{\it building floating shelves} and {\it making latte}.
We find 65 related tasks by finding related tasks for each
primary task. We generate potential related tasks for a primary task by comparing the wikiHow
articles using a TF-IDF on a bag-of-words representation, which finds tasks with similar descriptions.
We then filter out near duplicates (e.g., {\it how to jack up a car} and {\it how to use a car jack}) by comparing top YouTube search results and removing candidates with overlaps, and manually remove a handful of irrelevant tasks.

We define steps and their order for each task by examining the
wikiHow articles, beginning with the summaries of each step. Using the wikiHow
summary itself is insufficient, since many articles contain non-visual steps and some steps combine multiple
physical actions. We thus manually correct the list yielding a set of tasks with 7.4 steps on average for primary tasks and 8.8 for related tasks.

We then obtain videos for each task by searching YouTube. Since the related
tasks are only to aid the primary tasks, we take the top 30 results from
YouTube. For primary tasks, we ask annotators to filter a larger pool of top results while examining
the video, steps, and wikiHow illustrations, yielding at least 80 videos per task.

\subsection{Annotations and Statistics}

\noindent {\bf Task localization annotations.}
Since our focus is the primary tasks, annotators mark the temporal extent of each primary task step
independently. We do this for our 18 primary tasks and make annotations publically available\footref{fn:dataset}.

\noindent {\bf Dataset.} This results in a dataset containing 2750 videos of 18 primary
tasks comprising 212 hours of video; and 1950 videos of 65 related tasks comprising
161 hours of video.  We contrast this dataset with past
instructional video datasets in Table~\ref{tab:datatable}. Our dataset is
simultaneously large while also having precise temporal segment annotations.

To illustrate the dataset, we report a few summary statistics about the
primary task videos.
The videos are quite long, with an average length of 4min 57sec, and depict
fairly complex tasks, with 7.4 steps on average.  Less complex tasks include
{\it jack up a car} (3 steps); more complex ones include {\it pickle cucumbers} or
{\it change tire} (11 steps each).

\noindent {\bf Challenges.}
In addition to being long and complex, these videos are challenging since they do
not precisely show the ordered steps we have defined.
For instance, in {\it add oil to car}, 85\% of frames
instead depict background information such as shots of people talking
or other things. This is not an outlier: on average $72\%$ of the dataset
is background. On the other hand, on average $31\%$ of steps are
not depicted due to variances in procedures and omissions ({\it pickle cucumber} has $48\%$ of steps missing).
Moreover, the steps do not necessarily appear in the correct order: to estimate the order consistency,
we compute an upper bound on performance using our given order and found that
the best order-respecting parse of the data still missed $14\%$ of steps.

\section{Experiments}
\label{sec:experiments}

\begin{table*}[t]
    \caption{Weakly supervised recall scores on test set (in \%). Our approach, which shares information across tasks, substantially
    and consistently outperforms non-sharing baselines. The standard deviation for reported scores does not exceed 1\%.}
\resizebox{\textwidth}{!}{
\begin{tabular}{lc@{~~~~}c@{~~}c@{~~}c@{~~}c@{~~}c@{~~}c@{~~}c@{~~}c@{~~}c@{~~}c@{~~}c@{~~}c@{~~}c@{~~}c@{~~}c@{~~}c@{~~}c@{~~}|c} \toprule
    & \rotatebox{90}{\small Make} \rotatebox{90}{\small Kimchi Rice}
    & \rotatebox{90}{\small Pickle} \rotatebox{90}{\small Cucumber}
    & \rotatebox{90}{\small Make Banana} \rotatebox{90}{\small Ice Cream}
    & \rotatebox{90}{\small Grill} \rotatebox{90}{\small Steak}
    & \rotatebox{90}{\small Jack Up } \rotatebox{90}{\small Car}
    & \rotatebox{90}{\small Make } \rotatebox{90}{\small Jello Shots}
    & \rotatebox{90}{\small Change } \rotatebox{90}{\small Tire}
    & \rotatebox{90}{\small Make } \rotatebox{90}{\small Lemonade}
    & \rotatebox{90}{\small Add Oil } \rotatebox{90}{\small to Car}
    & \rotatebox{90}{\small Make } \rotatebox{90}{\small Latte}
    & \rotatebox{90}{\small Build } \rotatebox{90}{\small Shelves}
    & \rotatebox{90}{\small Make } \rotatebox{90}{\small Taco Salad}
    & \rotatebox{90}{\small Make } \rotatebox{90}{\small French Toast}
    & \rotatebox{90}{\small Make } \rotatebox{90}{\small Irish Coffee}
    & \rotatebox{90}{\small Make } \rotatebox{90}{\small Strawberry Cake}
    & \rotatebox{90}{\small Make } \rotatebox{90}{\small Pancakes}
    & \rotatebox{90}{\small Make } \rotatebox{90}{\small Meringue}
    & \rotatebox{90}{\small Make } \rotatebox{90}{\small Fish Curry}
    & \rotatebox{90}{\small Average }
\\ \midrule
Supervised                           & 19.1          & 25.3          & 38.0          & 37.5          & 25.7          & 28.2          & 54.3          & 25.8          & 18.3         & 31.2          & 47.7          & 12.0          & 39.5          & 23.4          & 30.9          & 41.1          & 53.4          & 17.3          & 31.6 \\ \midrule
Uniform                              & 4.2           & 7.1           & 6.4           & 7.3           & \textbf{17.4} & 7.1           & 14.2          & 9.8           & 3.1          & 10.7          & 22.1          & 5.5           & 9.5           & 7.5           & 9.2           & 9.2           & 19.5          & 5.1           & 9.7 \\
Alayrac'16~\cite{Alayrac15Unsupervised} & \textbf{15.6} & 10.6          & 7.5           & 14.2          & 9.3           & 11.8          & 17.3          & 13.1          & \textbf{6.4} & 12.9          & 27.2          & 9.2           & 15.7          & 8.6           & 16.3          & 13.0          & 23.2          & 7.4           & 13.3 \\
Richard'18~\cite{richard18actionsets} & 7.6 & 4.3 & 3.6 & 4.6 & 8.9 & 5.4 & 7.5 & 7.3 & 3.6 & 6.2 & 12.3 & 3.8 & 7.4 & 7.2 & 6.7 & 9.6 & 12.3 & 3.1 & 6.7 \\
Task-Specific Step-Based &     13.2      & 17.6 & 19.3          & 19.3          & 9.7           & 12.6          & 30.4          & 16.0          & 4.5          & 19.0          & 29.0 & 9.1           & 29.1          & \textbf{14.5}          & 22.9          & 29.0          & 32.9          & 7.3           & 18.6 \\
Proposed                             & 13.3          & \textbf{18.0}          & \textbf{23.4} & \textbf{23.1} & 16.9          & \textbf{16.5} & \textbf{30.7} & \textbf{21.6} & 4.6          & \textbf{19.5} & \textbf{35.3}          & \textbf{10.0} & \textbf{32.3} & 13.8 & \textbf{29.5} & \textbf{37.6} & \textbf{43.0} & \textbf{13.3} & \textbf{22.4} \\ \midrule
Gain from Sharing                    & 0.2           & 0.4          & 4.1           & 3.8           & 7.2           & 3.9           & 0.3           & 5.6           & 0.1         & 0.6           & 6.3          & 0.9           & 3.2           & -0.7           & 6.6           & 8.7           & 10.1           & 6.0           & 3.7 \\
\bottomrule
\end{tabular}
}
\vspace{-0.1in}
\label{tab:tab_results}
\end{table*}

Our experiments aim to address the following three questions about cross-task sharing in the weakly-supervised setting:
{\bf (1)} Can the proposed method use related data to improve performance?
{\bf (2)} How does the proposed component model compare to sharing alternatives?
{\bf (3)} Can the component model transfer to previously unseen tasks?
Throughout, we evaluate on the large dataset introduced in Section \ref{sec:dataset} that consists of
primary tasks and related tasks.
We address (1) in Section \ref{subsec:exp_multitask} by comparing our proposed approach with methods that do not share
and show that our proposed approach can use related tasks to improve performance on primary asks.
Section \ref{subsec:exp_sharing} addresses (2) by analyzing the performance of the model and showing that it outperforms step-based
alternatives.
We answer (3) empirically in Section \ref{subsec:exp_transfer} by training only on related tasks, and show that we are able to perform well on primary tasks.

\subsection{Cross-task Learning}
\label{subsec:exp_multitask}

We begin by evaluating whether our proposed component model approach can use sharing to improve performance on a fixed set of tasks.
We fix our evaluation to be the 18 primary tasks and evaluate whether the model can use the 65
related tasks to improve performance.

\noindent {\bf Metrics and setup.} We evaluate results on 18 primary tasks over the videos that
make up the test set. We quantify performance via {\it recall}, which we
define as the ratio between the number of correct step assignments (defined as
falling into the correct ground-truth time interval) and the total
number of steps over all videos. In other words, to get a perfect score, a method
must correctly identify one instance of each step of the task in each test video.
All methods make a single prediction per step, which prevents the trivial solution of assigning all frames to all actions.

We run experiments 20 times, each time making a train set of 30 videos per
task and leaving the remaining 1850 videos for test. We report the average.
Hyperparameters are set for all methods using a fixed validation set of 20
videos per primary task that are never used for training or testing.

\begin{figure}[t]
\begin{center}
    \includegraphics[width=\linewidth]{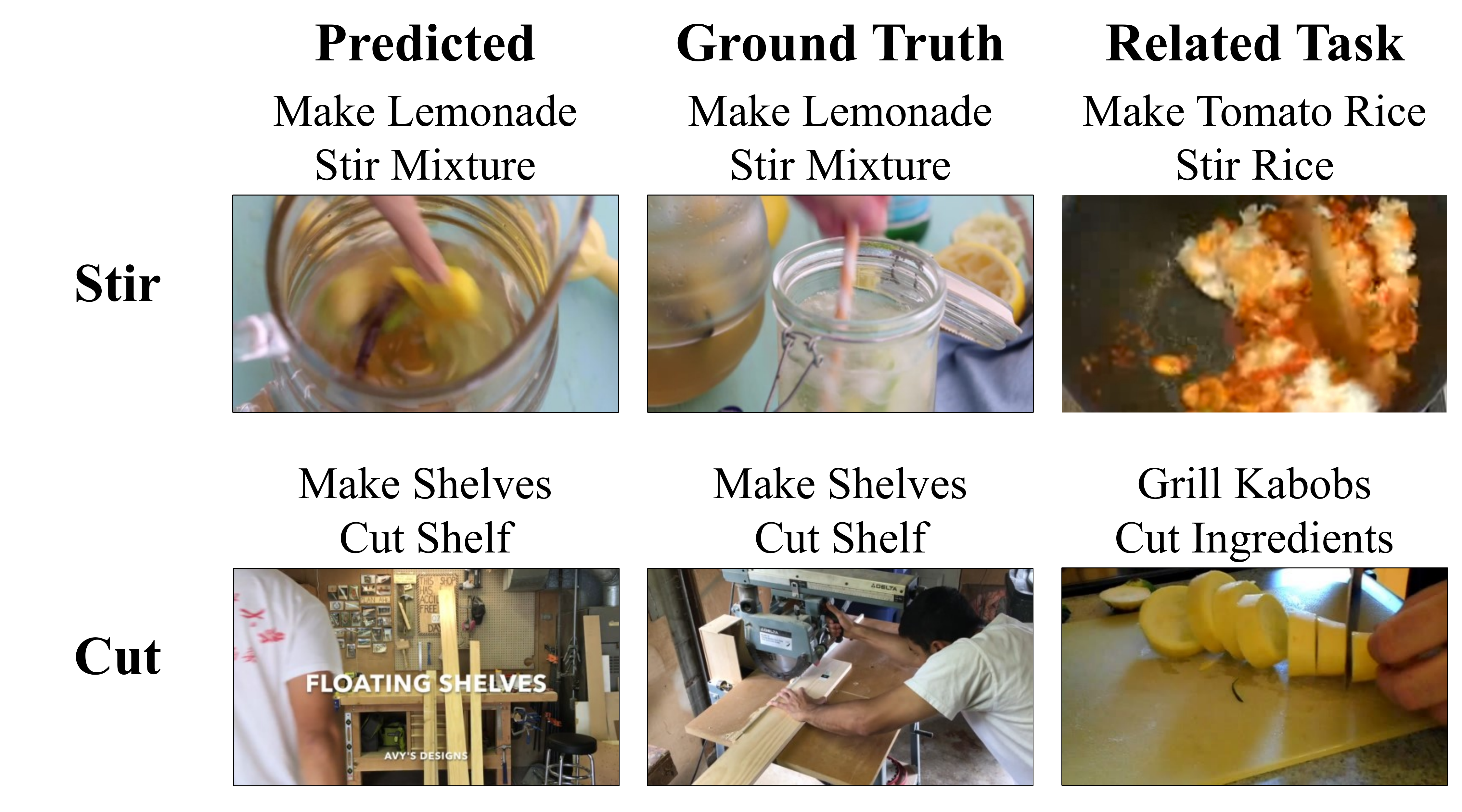}
   \caption{Components that share well and poorly: while stir shares well between steps of tasks, cut shares poorly when
       transferring from a food context to a home improvement context.}
   \label{fig:qualitative_share}
\end{center}
\vspace{-0.2in}
\end{figure}

\noindent {\bf Baselines.} Our goal is to examine whether our
sharing approach can leverage related tasks to improve performance on our
primary task. We compare our method to its version without sharing as well as to a number of baselines.
{\it (1)~Uniform:} simply predict steps at fixed time intervals. Since this predicts steps in the correct order and steps often break tasks into roughly equal chunks, this is fairly well-informed prior.
{\it (2)~Alayrac'16:} the weakly supervised learning method for videos, proposed in \cite{Alayrac15Unsupervised}. This
is similar in spirit to our approach except it does not share and optimizes a
L2-criterion via the DIFFRAC \cite{Bach07diffrac} method.
{\it (3)~Richard'18:} the weakly supervised learning method~\cite{richard18actionsets} that does not rely on the known order of steps.
{\it (4)~Task-Specific Steps:}
Our approach trained independently for each step of each task. In other words,
there are separate models for {\it pour egg} in the contexts of {\it making pancakes}
and {\it making meringue}. This differs from Alayrac in that it optimizes a cross-entropy loss using our proposed
optimization method. It differs from our full proposed approach since it
performs no sharing.
Note, that the full method in \cite{Alayrac15Unsupervised} includes automatic discovery of steps from narrations. Here, we only use the visual model of \cite{Alayrac15Unsupervised}, while providing the same constraints as in our method. This allows for a fair comparison between \cite{Alayrac15Unsupervised} and our method, since both use the same amount of supervision.
At test time, the method presented in \cite{richard18actionsets} has no prior about which steps are present or the order in which they occur. To make a fair comparison, we use the trained classifier of the method in \cite{richard18actionsets}, and apply the same inference procedure as in our method.

\noindent {\bf Qualitative results.} We illustrate qualitative results of our full method
in Figure~\ref{fig:exp_qualitative}. We show a parses of unseen videos of {\it Build Shelves} and
{\it Make Banana Ice Cream} and failure modes.
Our method can handle well a large variety of tasks and steps
but may struggle to identify some details (e.g., vanilla vs. egg) or actions.

\noindent {\bf Quantitative results.}
Table \ref{tab:tab_results} shows results summarized across steps.
The uniform baseline provides a strong lower bound, achieving an average recall of 9.7\% and outperforming \cite{richard18actionsets}. Note, however, that \cite{richard18actionsets} is designed to adress a different problem and cannot be fairly compared with other methods in our setup.
While \cite{Alayrac15Unsupervised} improves on this (13.3\%), it does substantially worse than our task-specific step method (18.6\%).
We found that predictions from \cite{Alayrac15Unsupervised} often had several steps with similar scores, leading to poor parse results, which we attribute to the convex relaxation used by DIFFRAC.
This was resolved in the past by the use of narration at test time; our approach does not depend on this.

Our full approach, which shares across tasks, produces substantially better performance (22.4\%) than the task-specific
step method. More importantly, this improvement is systematic: the full method improves on the task-specific step
baseline in 17 tasks out of 18.

We illustrate some qualitative examples of steps benefiting and least benefiting from sharing
in Figure~\ref{fig:qualitative_share}. Typically, sharing can help if the component has distinctive
appearance and is involved in a number of steps: steps involve stirring, for
instance, have an average gain of 15\% recall over independent training because it is frequent (in 30 steps)
and distinctive. Of course, not all steps benefit: {\it cut shelf} is harmed ($47\%$ independent $\to$ $28\%$ shared)
because {\it cut} mostly occurs in cooking tasks with dissimilar contexts.

\noindent {\bf Verifying optimizer on small-scale data.} We now evaluate our approach on the smaller 5-task
dataset of \cite{Alayrac15Unsupervised}.  Since here there are no common steps across tasks, we are able to test only the
basic task-specific step-based version.  To make a fair comparison, we use the same features, ordering constraints, as
well as constraints from narration for every K as provided by the authors of \cite{Alayrac15Unsupervised}, and we
evaluate using the F1 metric as in \cite{Alayrac15Unsupervised}.  As a result, the two formulations are on par, where
\cite{Alayrac15Unsupervised} versus our approach result in 22.8\% versus 21.8\% for K=10 and 21.0\% versus 21.1\% for
K=15, respectively. While these scores are slightly lower compared to those obtained by the single-task probabilistic model
in Sener \cite{sener18unsupervised} (25.4\% at K=10 and 23.6\% at K=15), we are unable to compare using our full cross-task model on this
dataset. Overall, these results verify the effectiveness of our optimization technique.

\subsection{Experimental Evaluation of Cross-task Sharing}
\label{subsec:exp_sharing}

Having verified the framework and the role of sharing, we now more precisely evaluate
how sharing is performed to examine the contribution of our proposed compositional model.
We vary two dimensions.  The first is the granularity, or at what level sharing occurs. We propose
sharing at a component level, but one could share at a step level as well. The second is
what data is used, including (i) independently learning primary tasks; (ii) learning primary
tasks together; (iii) learning primary plus related tasks together.

Table~\ref{tab:step_vs_attr} reveals that increased sharing consistently helps and component-based
sharing extracts more from sharing than step-based (performance increases across rows). This gain over
step-based sharing is because step-based sharing requires exact matches.
Most commonality between tasks occurs with slight variants (e.g., {\it cut} is
applied to steak, tomato, pickle, etc.) and therefore a component-based
model is needed to maximally enable sharing.

\begin{table}
\caption{
Average recall scores on the test set for our method when changing the sharing settings and the model.
}
\label{tab:step_vs_attr}
\begin{tabular}{lccc} \toprule
     & \small Unshared & \small Shared & \small Shared Primary \\
     & \small Primary & \small Primary & \small + Related \
     \\ \midrule
Step-based      & 18.6          & 18.9          & 19.8 \\
Component-based & \textbf{18.7} & \textbf{20.2} & \textbf{22.4} \\
\bottomrule
\end{tabular}
\vspace{-0.1in}
\end{table}

\begin{figure}[t]
\begin{center}
   \includegraphics[width=\linewidth]{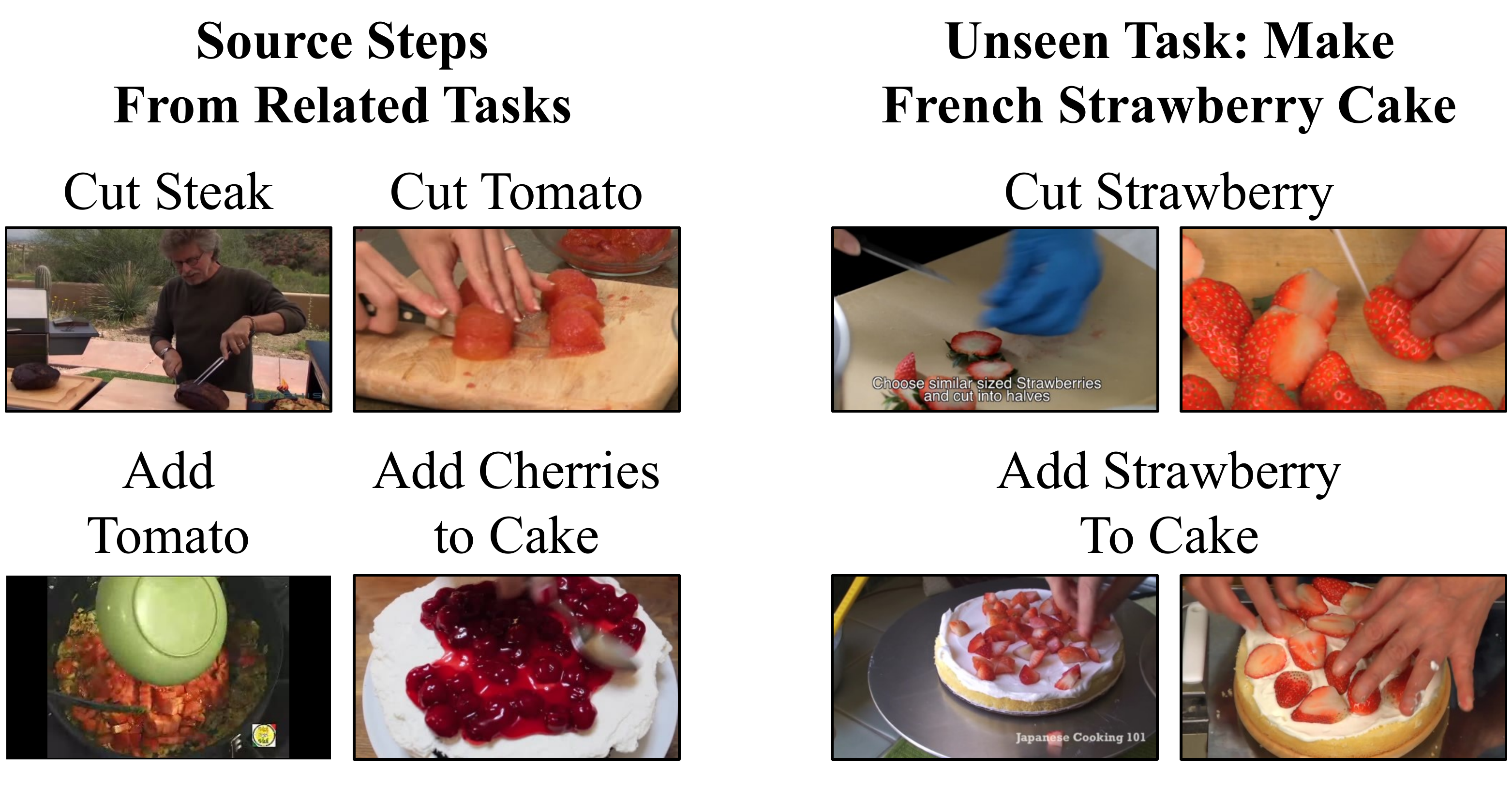}
   \caption{Examples of identified steps for an unseen task. While the model has not seen these steps
and objects e.g.,~strawberries, its knowledge of other components leads to reasonable predictions.}
   \label{fig:transferred}
\end{center}
\vspace{-0.2in}
\end{figure}

\subsection{Novel Task Transfer}
\label{subsec:exp_transfer}

One advantage of shared representations is that they can let one parse
new concepts. For example, without any modifications, we can repeat our
experiments from Section~\ref{subsec:exp_multitask} in a setting where we
never train on the 18 tasks that we test on but instead on
the 65 related tasks.
The only information given about the test tasks is an ordered list of steps.

\noindent {\bf Setup.} As in Section~\ref{subsec:exp_multitask},
we quantify performance with recall on the 18 primary tasks. However,
we train on a subset of the 65 related tasks and never on any primary
task.

\begin{figure}[t]
\centering
\includegraphics[width=0.9\linewidth]{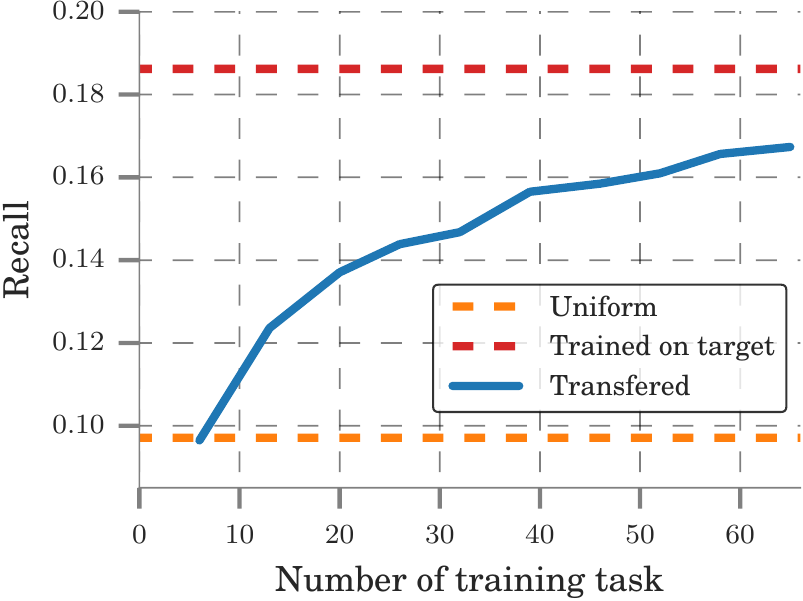}
\caption{Recall while transferring a learned model to unseen tasks as a function of the number
of tasks used for training. Our component model approaches training directly on these tasks. }
\label{fig:res_transfer}
\end{figure}

\noindent {\bf Qualitative results.} We show a parse of steps of {\it Make Strawberry Cake}
in Figure~\ref{fig:transferred} using all related tasks. The model has not seen {\it cut strawberry} before
but has seen other forms of cutting. Similarly, it has seen {\it add cherries
to cake}, and can use this step to parse {\it add strawberries to cake}.

\noindent {\bf Quantitative results.} Figure~\ref{fig:res_transfer} shows performance as a function
of the number of related tasks used for training. Increasing the number of training tasks
improves performance on the primary tasks, and does not plateau
even when 65 tasks are used.

\section{Conclusion}
\label{sec:conclusions}

We have introduced an approach for weakly supervised learning
from instructional videos and a new CrossTask dataset for evaluating the role of sharing in this
setting. Our component model has been shown ability to exploit common parts of tasks
to improve performance and was able to parse previously unseen tasks.
Future work
would benefit from improved features as well as from improved versions of sharing.

\paragraph{Acknowledgements.} This work was supported in part by the MSR-Inria joint lab, the Louis Vuitton ENS Chair on Artificial Intelligence, ERC grants LEAP No.~336845 and ACTIVIA No.~307574, the DGA project DRAAF, CIFAR Learning in Machines \& Brains program, the European Regional Development Fund under the project IMPACT (reg. no. CZ.02.1.01/0.0/0.0/15003/0000468), the TUBITAK Grant 116E445 and a research fellowship by the Embassy of France.
We thank Francis Bach for helpful discussions about the optimization procedure.

{\small
\bibliographystyle{ieee}
\bibliography{biblio}

\begin{thebibliography}{10}\itemsep=-1pt

\bibitem{Alayrac15Unsupervised}
J.-B. Alayrac, P.~Bojanowski, N.~Agrawal, I.~Laptev, J.~Sivic, and
  S.~Lacoste~Julien.
\newblock Unsupervised learning from narrated instruction videos.
\newblock In {\em CVPR}, 2016.

\bibitem{alayrac16objectstates}
J.-B. Alayrac, J.~Sivic, I.~Laptev, and S.~Lacoste-Julien.
\newblock Joint discovery of object states and manipulation actions.
\newblock In {\em ICCV}, 2017.

\bibitem{Bach07diffrac}
F.~Bach and Z.~Harchaoui.
\newblock {DIFFRAC}: {A} discriminative and flexible framework for clustering.
\newblock In {\em NIPS}, 2007.

\bibitem{bojanowski17unsupervised}
P.~Bojanowski and A.~Joulin.
\newblock Unsupervised learning by predicting noise.
\newblock In {\em ICML}, 2017.

\bibitem{Bojanowski14weakly}
P.~Bojanowski, R.~Lajugie, F.~Bach, I.~Laptev, J.~Ponce, C.~Schmid, and
  J.~Sivic.
\newblock Weakly supervised action labeling in videos under ordering
  constraints.
\newblock In {\em ECCV}, 2014.

\bibitem{Bojanowski15weakly}
P.~Bojanowski, R.~Lajugie, E.~Grave, F.~Bach, I.~Laptev, J.~Ponce, and
  C.~Schmid.
\newblock Weakly-supervised alignment of video with text.
\newblock In {\em ICCV}, 2015.

\bibitem{Caron2018}
M.~Caron, P.~Bojanowski, A.~Joulin, and M.~Douze.
\newblock {Deep Clustering for Unsupervised Learning of Visual Features}.
\newblock In {\em ICCV}, 2018.

\bibitem{carreira17quovadis}
J.~Carreira and A.~Zisserman.
\newblock Quo vadis, action recognition? a new model and the kinetics dataset.
\newblock In {\em CVPR}, 2017.

\bibitem{Damen2018}
D.~Damen, H.~Doughty, G.~Maria~Farinella, S.~Fidler, A.~Furnari, E.~Kazakos,
  D.~Moltisanti, J.~Munro, T.~Perrett, W.~Price, and M.~Wray.
\newblock Scaling egocentric vision: The {EPIC-KITCHENS} dataset.
\newblock In {\em ECCV}, 2018.

\bibitem{dima2014youdo}
D.~Damen, T.~Leelasawassuk, O.~Haines, A.~Calway, and W.~Mayol-Cuevas.
\newblock You-do, i-learn: Discovering task relevant objects and their modes of
  interaction from multi-user egocentric video.
\newblock In {\em BMVA}, 2014.

\bibitem{Fang2018}
K.~Fang, T.-L. Wu, D.~Yang, S.~Savarese, and J.~J. Lim.
\newblock Demo2vec: Reasoning object affordances from online videos.
\newblock In {\em CVPR}, 2018.

\bibitem{Farhadi09}
A.~Farhadi, I.~Endres, D.~Hoiem, and D.~Forsyth.
\newblock Describing objects by their attributes.
\newblock In {\em CVPR}, 2009.

\bibitem{Ferrari07}
V.~Ferrari and A.~Zisserman.
\newblock Learning visual attributes.
\newblock In {\em NIPS}, 2007.

\bibitem{Fouhey18}
D.~F. Fouhey, W.~Kuo, A.~A. Efros, and J.~Malik.
\newblock From lifestyle vlogs to everyday interactions.
\newblock In {\em CVPR}, 2018.

\bibitem{guadarrama13}
S.~Guadarrama, N.~Krishnamoorthy, G.~Malkarnenkar, S.~Venugopalan, R.~Mooney,
  T.~Darrell, and K.~Saenko.
\newblock Youtube2text: Recognizing and describing arbitrary activities using
  semantic hierarchies and zero-shot recognition.
\newblock In {\em ICCV}, 2013.

\bibitem{he16resnet}
K.~He, X.~Zhang, S.~Ren, and J.~Sun.
\newblock Deep residual learning for image recognition.
\newblock In {\em CVPR}, 2016.

\bibitem{hershey17audio}
S.~Hershey, S.~Chaudhuri, D.~P.~W. Ellis, J.~F. Gemmeke, A.~Jansen, C.~Moore,
  M.~Plakal, D.~Platt, R.~A. Saurous, B.~Seybold, M.~Slaney, R.~Weiss, , and
  K.~Wilson.
\newblock Cnn architectures for large-scale audio classification.
\newblock In {\em ICASSP}, 2017.

\bibitem{feifei2016connectionist}
D.-A. Huang, L.~Fei-Fei, and J.~C. Niebles.
\newblock Connectionist temporal modeling for weakly supervised action
  labeling.
\newblock In {\em ECCV}, 2016.

\bibitem{huang17unsupervised}
D.-A. Huang, J.~J. Lim, L.~Fei-Fei, and J.~C. Niebles.
\newblock Unsupervised visual-linguistic reference resolution in instructional
  videos.
\newblock In {\em CVPR}, 2017.

\bibitem{huang18finding}
D.-A. Huang, V.~Ramanathan, D.~Mahajan, L.~Torresani, M.~Paluri, L.~Fei-Fei,
  and J.~C. Niebles.
\newblock Finding "it": Weakly-supervised reference-aware visual grounding in
  instructional video.
\newblock In {\em CVPR}, 2018.

\bibitem{adam}
D.~Kingma and J.~Ba.
\newblock Adam: A method for stochastic optimization.
\newblock {\em arXiv preprint arXiv:1412.6980}, 2014.

\bibitem{kuehne17weakly}
H.~Kuehne, A.~Richard, and J.~Gall.
\newblock Weakly supervised learning of actions from transcripts.
\newblock In {\em CVIU}, 2017.

\bibitem{Liu11}
J.~Liu, B.~Kuipers, and S.~Savarese.
\newblock Recognizing human actions by attributes.
\newblock In {\em CVPR}, 2011.

\bibitem{Malmaud15what}
J.~Malmaud, J.~Huang, V.~Rathod, N.~Johnston, A.~Rabinovich, and K.~Murphy.
\newblock What's cookin'? {I}nterpreting cooking videos using text, speech and
  vision.
\newblock In {\em NAACL}, 2015.

\bibitem{misra2017composing}
I.~Misra, A.~Gupta, and M.~Hebert.
\newblock {From Red Wine to Red Tomato: Composition with Context}.
\newblock In {\em CVPR}, 2017.

\bibitem{richard17weakly}
A.~Richard, H.~Kuehne, and J.~Gall.
\newblock Weakly supervised action learning with rnn based fine-to-coarse
  modeling.
\newblock In {\em CVPR}, 2017.

\bibitem{richard18actionsets}
A.~Richard, H.~Kuehne, and J.~Gall.
\newblock Action sets: Weakly supervised action segmentation without ordering
  constraints.
\newblock In {\em CVPR}, 2018.

\bibitem{sener18unsupervised}
F.~Sener and A.~Yao.
\newblock Unsupervised learning and segmentation of complex activities from
  video.
\newblock In {\em CVPR}, 2018.

\bibitem{Sener15unsupervised}
O.~Sener, A.~Zamir, S.~Savarese, and A.~Saxena.
\newblock Unsupervised semantic parsing of video collections.
\newblock In {\em ICCV}, 2015.

\bibitem{simonyan2014two}
K.~Simonyan and A.~Zisserman.
\newblock Two-stream convolutional networks for action recognition in videos.
\newblock In {\em NIPS}, 2014.

\bibitem{Wang13action}
H.~Wang and C.~Schmid.
\newblock Action recognition with improved trajectories.
\newblock In {\em ICCV}, 2013.

\bibitem{Xu2004maximum}
L.~Xu, J.~Neufeld, B.~Larson, and D.~Schuurmans.
\newblock Maximum margin clustering.
\newblock In {\em NIPS}, 2004.

\bibitem{Yao11}
B.~Yao, X.~Jiang, A.~Khosla, A.~L. Lin, L.~Guibas, and L.~Fei-Fei1.
\newblock Human action recognition by learning bases of action attributes and
  parts.
\newblock In {\em ICCV}, 2011.

\bibitem{Yatskar_Commonly_17}
M.~Yatskar, V.~Ordonez, L.~Zettlemoyer, and A.~Farhadi.
\newblock Commonly uncommon: Semantic sparsity in situation recognition.
\newblock In {\em Proceedings of the CVPR}, 2017.

\bibitem{zhou18towards}
L.~Zhou, X.~Chenliang, and J.~J. Corso.
\newblock Towards automatic learning of procedures from web instructional
  videos.
\newblock In {\em AAAI}, 2018.

\end{thebibliography}


\begin{thebibliography}{1}\itemsep=-1pt

\bibitem{alayrac:hal-01580630}
J.-B. Alayrac, P.~Bojanowski, N.~Agrawal, J.~Sivic, I.~Laptev, and
  S.~Lacoste-Julien.
\newblock {Learning from narrated instruction videos}.
\newblock {\em {IEEE Transactions on Pattern Analysis and Machine
  Intelligence}}, XX, Sept. 2017.

\bibitem{NIPS2013_5021}
T.~Mikolov, I.~Sutskever, K.~Chen, G.~S. Corrado, and J.~Dean.
\newblock Distributed representations of words and phrases and their
  compositionality.
\newblock In C.~J.~C. Burges, L.~Bottou, M.~Welling, Z.~Ghahramani, and K.~Q.
  Weinberger, editors, {\em Advances in Neural Information Processing Systems
  26}, pages 3111--3119. Curran Associates, Inc., 2013.

\bibitem{bojanowski17wv}
A.~J. Piotr~Bojanowski, Edouard~Grave and T.~Mikolov.
\newblock Enriching word vectors with subword information.
\newblock {\em arXiv}, 2017.

\bibitem{maaten08tsne}
L.~van~der Maaten and G.~Hinton.
\newblock Visualizing data using t-sne.
\newblock {\em Journal of machine learning research}, 2008.

\end{thebibliography}
}

\clearpage

\appendix

\setcounter{section}{0}
\renewcommand{\thesection}{\Alph{section}}
\setcounter{table}{0}
\setcounter{figure}{0}

\section{Outline of supplementary material}
This supplementary material provides more details on our method and our dataset, together with some additional results of our method.

In Section \ref{sec:model} we describe details of our narration-based temporal constraints and the optimization procedure.
Section \ref{sec:dataset} provides more information about our dataset and video collection procedure, including a complete list of primary and related tasks and task-wise statistics.
In Section \ref{sec:experiments} we illustrate additional quantitative and qualitative results of our method, including classifier scores and localized steps.
We also provide more examples and analysis of failure cases.

\section{Modeling instructional videos}
\label{sec:model}
\subsection{Temporal text localization}
In this section we explain in detail how we obtain temporal constraints from subtitles of the video.
We assume that each step in the video occurs roughly at the same time as it is mentioned in the narration.
Step localization in narrations is challenging for several reasons.
First, the same step may be described in different ways (\eg \textit{cut steak} and \textit{slice meat}).
It may contain a reference (\eg \textit{cut it}).
Second, a mention of a step doesn't guarantee, that the step occurs at the same time (\eg Let the steak rest before \textit{cutting it}).
Since most of the videos in our dataset are unprofessional, the narrator doesn't usually follow a strict scenario and often talks about unrelated topics.
Finally, most of the subtitles are produced by YouTube automatic speech recognition, and, therefore, contain errors and lack punctuation.

As described in Section \ref{subsec:video_collection} of the main paper, we provide a short textual description of each step.
These descriptions are matched to the text within a sliding window over the subtitles, in order to find where each step is mentioned.
More formally, let $f$ be a function, mapping a sequence of words of variable length into $\mathbb{R}^D$.
Applying this function to the text within a sliding window of size $w$ yields a matrix $U\in\mathbb{R}^{L\times D}$, where $L$ is the number of words in the subtitles.
Applying the same function to the description of each step gives us a matrix $V\in\mathbb{R}^{K\times D}$, where $K$ is a total number of steps.
Assuming that $\sum\limits_{d=1}^DU_{ld}^2=1$ for any $l=1\dots L$, and that $\sum\limits_{d=1}^DV_{kd}^2=1$ for any $k=1\dots K$
(i.e., the features are unit-norm), $S=UV^T\in\mathbb{R}^{L\times K}$ is a matrix of cosine similarities between vector representations of subtitles and descriptions of steps.

We find the best matching $A\in\{0,1\}^{L\times K}$ between the steps and the subtitles, that satisfies the ordering of the steps, by solving a linear problem
\begin{equation}
	\label{eq:text_consts}
	\min_{A\in\mathcal{A}}\sum\limits_{l,k}S_{l,k}A_{l,k}
\end{equation}
where $\mathcal{A}$ is a set of assignments that satisfy at-least-one and ordering constraints.
This problem can be efficiently solved via dynamic programming, as described in Section \ref{subsec:dl}.
Imposing the ordering constraints during step localization helps to avoid spurious mentions of steps, that don't follow the scenario of a task.

We try different choices of mapping $f$.
The first is TF-IDF representation of the text within sliding window.
The second is a Word2Vec-like word embedding \citesup{NIPS2013_5021}, followed by a max-pooling over sliding window.
We obtain our word embedding by training the Fasttext model \citesup{bojanowski17wv} with dimension $100$.
The model is trained on a corpus of subtitles of 2 million YouTube videos for 6729 tasks from wikiHow.

Finally, we propose a way to learn a better aggregation function than max-pooling for the word vectors.
Assume that we are given a set of sentences $\mathcal{I}$ of various lengths.
Each sentence $i$ is represented by $X^i\in\mathbb{R}^{M_i \times d_0}$, where $M_i$ is the number of words in a sentence and $X^i_m\in\mathbb{R}^{d_0}$ is the feature vector corresponding to $m$-th word in the sentence.
Assume that for each sentence $i$ we are also given a set of sentences $S_i\subset\mathcal{I}$ with similar meaning and a set of sentences $D_i\subset\mathcal{I}$ with different meaning.
$f$ is learnt by minimizing loss
\begin{equation}
	\label{eq:textloss}
	\begin{aligned}
		\sum\limits_{i\in\mathcal{I}}\frac1{|S_i|}\sum\limits_{j\in S_i, k\in D_i}\max[0, \cossim(f(X^i),f(X^k)) \\ - \cossim(f(X^i),f(X^j)) + h],
	\end{aligned}
\end{equation}
where $\cossim(a,b)=\frac{a^\text{T} b}{\left\lVert a\right\rVert\left\lVert b\right\rVert}$ is the cosine similarity function, and, $h$ is the margin constant.
This can be understood as pushing representations of sentences with similar meaning closer together, as opposed to sentences with different meaning.
We take $f$ in the form of 1D convolution with kernel length $1$ and a number of filters $d$, followed by the global max-pooling, and by a linear mapping $\mathbb{R}^d\to\mathbb{R}^d$. We take $d=300$ and $h=0.1$.

\begin{figure*}[t]
\begin{center}
   \includegraphics[width=\linewidth]{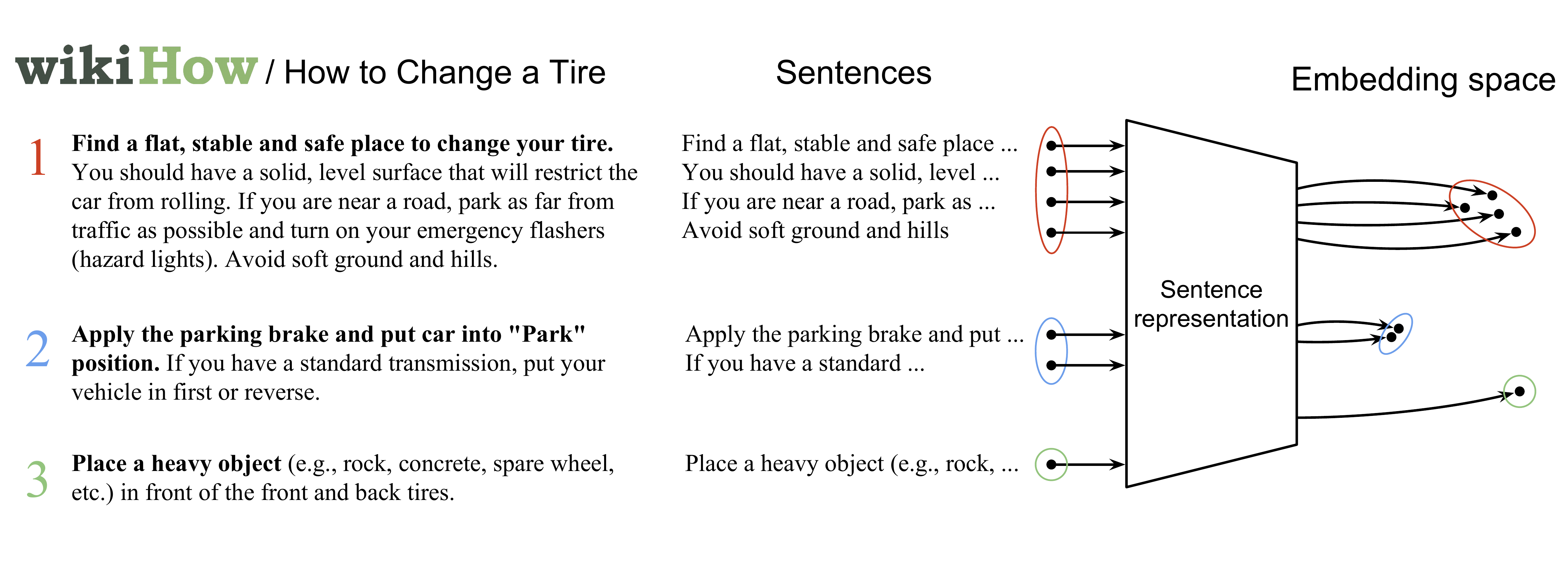}
   \caption{Example of three wikiHow steps for the \textit{Change a Tire} task. Our method learns a similarity function that pulls representations of the sentences from the same paragraphs closer together, and pushes the sentences from different paragraphs away from each other}
   \label{fig:wikisteps}
\end{center}
\vspace{-0.2in}
\end{figure*}

We train our model on the set of sentences from wikiHow.
Descriptions of the tasks on wikiHow are organized into step paragraphs, as shown on figure \ref{fig:wikisteps}.
We assume that sentences within the same paragraph describe similar concepts, while sentences from different steps of the same task have different meanings.
For each sentence $i$, $S_i$ is defined as a set of sentences from the same paragraph, and $D_i$ is defined as a set of sentences from other paragraphs within the same wikiHow page.

We evaluate alternative text representations by comparing obtained constraints with the ground truth on the set of primary tasks.
The results are shown in Table \ref{tab:constr_eval}.
Our aggregation function, trained on wikiHow outperforms TF-IDF and max-pooled word vectors both in terms of precision and recall.

\begin{table}[]
\caption{Precision and recall of the constraints obtained with our method, averaged over 18 primary tasks.}
\label{tab:constr_eval}
\begin{tabular}{@{}lcc@{}}
\toprule
 & Precision (\%) & Recall (\%) \\
\midrule
Max-pooled word vectors & 11.6 & 10.4 \\
TF-IDF & 13.3 & 11.4 \\
Proposed & \textbf{15.9} & \textbf{13.9} \\
\bottomrule
\end{tabular}
\vspace{-0.1in}
\end{table}

\subsection{Constrained linear optimization}
\label{subsec:dl}
Our optimization procedure, inference and temporal text localization require solving a linear problem of the form
\begin{equation}
	\label{eq:linear}
	\min_{Y\in\mathcal{C}}\sum\limits_{t,k}S_{tk}Y_{tk},
\end{equation}
where $S\in\mathbb{R}^{T\times K}$ and $\mathcal{C}$ is the set of all assignments from $\{0,1\}^{T\times K}$ that satisfy the {\em ordering} and {\em at-least-once} constraints.
At-least-once constraints mean that every step $k$ should be picked at least once: $\sum\limits_{t}Y_{t,k} \geq 1$ for any $k = 1\dots K$.
Ordering constraints mean that the step $k-1$ should precede step $k$ for any $k\geq 2$.
This problem can be solved efficiently via dynamic programming.
First, we rewrite the problem in the form:
\begin{equation}
	\min_{y\in\tilde{C}}\sum\limits_{t}S_{ty_t},
\end{equation}
where $y_t \in \{0,1,\dots,K\}$ is the step label at time $t$ ($0$ stands for background, i.e. when no step is selected).
The ordering constraints impose that $y_{t+1}\in \{0,z_t\}$, where $z_t = \max(y_1,\dots,y_t)$ is the last non-background step.
We define state $x_t$ at time $t$ as a pair $(y_t,z_t)$.
Note, that for a given state $x_t$, the only possible $x_{t-1}$ that satisfies the constraints are $(z_t,z_t)$, $(z_t-1,z_t-1)$ and $(0,z_t-1)$ if $y_t\neq 0$, and $(z_t-1,z_t-1)$ and $(0,z_t-1)$ otherwise.
We denote this set of possible previous states as $\mathcal{P}(x_t)$.
The minimum cumulative cost for state $x_t=x$ at time $t$ is
\begin{equation}
	\label{eq:state_cost}
	V(x,t) = \min_{x_1,\dots,x_{t-1}~|~x_t=x}\sum\limits_{\tau=1}^tS_{\tau y_\tau}.
\end{equation}

Define $C(x_t,x_{t-1})=S_{ty_t}$ if $x_{t-1}\in\mathcal{P}(x_t)$ and $C(x_t,x_{t-1})=+\infty$ otherwise (for simplicity we denote $x_0=(0,0)$). This allows to rewrite \eqref{eq:state_cost} in the recursive form:
\begin{equation}
	\label{eq:bellman}
	V(x,t) = \min_{x'}(C(x,x')+V(x',t-1)).
\end{equation}

We compute $V(x,t)$ recursively for $t=1,\dots,T$, using \eqref{eq:bellman}.
In practice, computing $V(x,t)$ given $V(x',t-1)$ for all $x'$ requires minimization only over $x'\in\mathcal{P}(x)$ and can be done in $\text{O}(1)$.
Since there are $2K$ possible states, the complexity of computing $V(x,t)$ for all $x$ and $t$ is $\text{O}(KT)$.
To satisfy {\em at-least-once} constraints, the final state $x_T$ must be either $(K,K)$, or $(0,K)$.
To get the optimal assignment, we take $x^*_T=\argmin\limits_{x\in\{(K,K),(0,K)\}}V(x,T)$ and find $x^*_t = \argmin\limits_{x\in\mathcal{P}(x^*_{t+1})}V(x,t-1)$ recursively for every $t=T-1,\dots,1$.

\begin{figure*}[t]
\begin{center}
   \includegraphics[width=\linewidth]{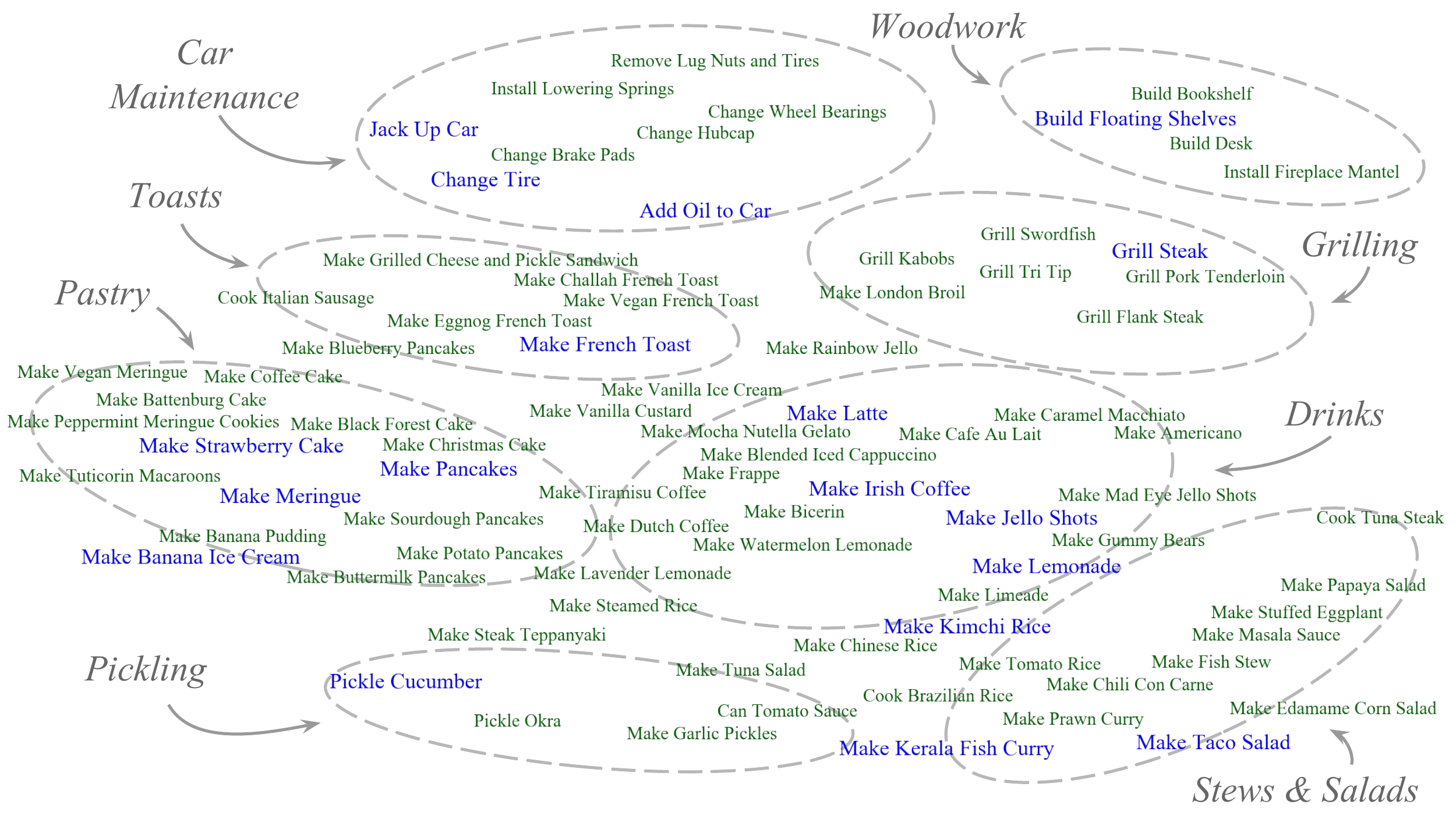}
   \caption{A t-SNE visualization of primary (in \textcolor{blue}{blue}) and related (in \textcolor{dark-green}{green}) tasks. The distance between two tasks is based on the number of components they share. Two well separable clusters on top correspond to {\em Car Maintenance} and {\em Home Repairs} categories, while most of the tasks belong to the {\em Cooking} category.}
   \label{fig:tasks}
\end{center}
\vspace{-0.2in}
\end{figure*}

\subsection{Optimization for discriminative clustering}
\label{subsec:dc_optim}
The discriminative clustering problem
\begin{equation}
	\label{eq:dc}
	\min_{Y\in\mathcal{C},F\in\mathcal{F}}-\sum\limits_{t,k}Y_{t,k}\log(\frac{\exp(f_k(x_t))}{\sum\limits_{k'}\exp(f_{k'}(x_t))}),
\end{equation}
introduced in Section 4.2 of the main paper can't be solved efficiently with
standard techniques, such as projected gradient descent, because the projection
over our constraint set $\mathcal{C}$ is computationally expensive.

Our optimization method can be applied to a broader class of problems of the form
\begin{equation}
	\label{eq:general_dc}
	\min_{Y\in\mathcal{C},\theta\in\mathbb{R}^m}\sum\limits_{tk}Y_{tk}F_{tk}(\theta).
\end{equation}

\noindent Given solution $(Y^l,\theta^l)$ at $l$-th iteration, we define a quadratic upper bound for $F(\theta)$ in the neighbourhood of $\theta^l$: $F_{tk}(\theta) \leq \tilde{F}_{tk}(\theta;\theta^l)$, where
\begin{equation}
	\label{eq:ubound}
	\tilde{F}_{tk}(\theta;\theta^l) = F_{tk}(\theta^l) + \nabla F_{tk}(\theta^l)(\theta-\theta^l) + \frac1{2\delta K}||\theta-\theta^l||^2.
\end{equation}

\noindent Note, that $\sum\limits_{t,k}Y_{tk}F_{tk}(\theta)\leq\sum\limits_{t,k}Y_{tk}\tilde{F}_{tk}(\theta;\theta^l)$ for any $(Y,\theta)$ and that $\sum\limits_{t,k}Y^l_{tk}F_{tk}(\theta^l) = \sum\limits_{t,k}Y^l_{tk}\tilde{F}_{tk}(\theta^l)$. This means, that for any $(Y^{l+1},\theta^{l+1})$, s.t. $\sum\limits_{t,k}Y^{l+1}_{tk}\tilde{F}_{tk}(\theta^{l+1}) \leq \sum\limits_{t,k}Y^l_{tk}\tilde{F}_{tk}(\theta^l)$, the same inequality holds for $F$:
\begin{equation}
	\sum\limits_{t,k}Y^{l+1}_{tk}F_{tk}(\theta^{l+1}) \leq \sum\limits_{t,k}Y^l_{tk}F_{tk}(\theta^l).
\end{equation}

\noindent For the problem
\begin{equation}
	\label{eq:min_ubound}
	\min_{Y\in\mathcal{C},\theta\in\mathbb{R}^m}\sum\limits_{t,k}Y_{tk}\tilde{F}_{tk}(\theta)
\end{equation}
it is possible to find a global minimum.
The minimization with respect to $\theta$ yields
\begin{equation}
	\theta^*(Y)	= \theta^l - \delta K\frac{\sum\limits_{t,k}Y_{tk}\nabla F_{tk}(\theta^l)}{\sum\limits_{t,k}Y_{tk}}.
\end{equation}

\noindent Since $\sum\limits_{t,k}Y_{tk} = K$, this expression can be simplified as
\begin{equation}
	\label{eq:opt_theta}
	\theta^*(Y)	= \theta^l - \delta\sum\limits_{t,k}Y_{tk}\nabla F_{tk}(\theta^l).
\end{equation}

\noindent Substituting $\theta$ with $\theta^*(Y)$ in \eqref{eq:min_ubound} leads to the problem
\begin{equation}
	\label{eq:linear_dc}
	\min_{Y\in\mathcal{C}}\sum_{t,k}[F_{tk}(\theta^l) - \frac{\delta}2||\nabla F_{tk}(\theta^l)||^2]Y_{tk}.
\end{equation}
This is a linear problem that can be solved as described in Section \ref{subsec:dl}.
Denote $Y^*$ a solution of \eqref{eq:linear_dc}. We obtain $\theta*$ by substituting $Y$s with $Y^*$ in \eqref{eq:opt_theta}. The pair $(Y^*,\theta^*)$ is a global minimum for \eqref{eq:min_ubound}. We take this pair as a new solution $(Y^{l+1},\theta^{l+1})$.

The described optimization procedure can be seen as alternating between solving a linear problem \eqref{eq:linear_dc} for $Y$ and a gradient descent step with the learning rate $\delta$
\begin{equation}
	\label{eq:grad_step}
	\theta^{l+1} = \theta^l - \delta\sum\limits_{t,k}Y^{l+1}_{tk}\nabla F_{tk}(\theta^l).
\end{equation}

\begin{figure}[t]
\begin{center}
   \includegraphics[width=.8\linewidth]{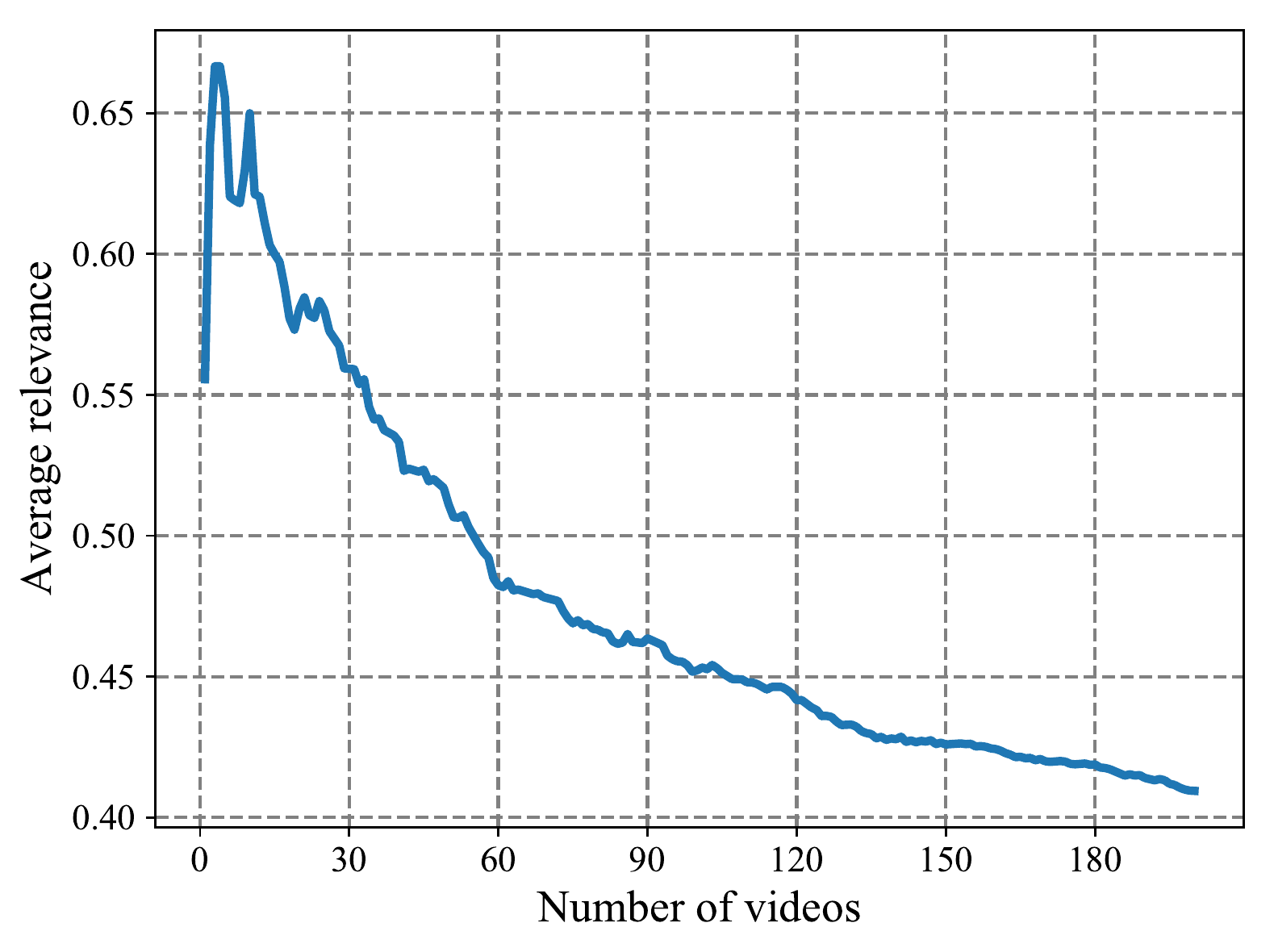}
   \caption{Average relevance of videos as a function of the number of videos collected from YouTube. Taking top 30 videos per task results in 56\% relevant videos. Attempting to collect more videos results in a noisy dataset with many irrelevant videos.}
   \label{fig:relevance}
\end{center}
\vspace{-0.2in}
\end{figure}

\begin{table*}[t]
\caption{Statistics for primary tasks.}
\label{tab:datastat}
\resizebox{\textwidth}{!}{
\begin{tabular}{@{}lcccccc@{}}
\toprule
Task                  & Number of videos & Number of steps & Average length & Missing steps & Background    & Order consistency \\ \midrule
Make Kimchi Rice      & 120              & 6               & 4:47           & 21\%          & 70\%          & 0,69              \\
Pickle Cucumber       & 106              & 11              & 5:35           & 48\%          & 75\%          & 0,85              \\
Make Banana Ice Cream & 170              & 5               & 4:04           & 38\%          & 80\%          & 0,98              \\
Grill Steak           & 228              & 11              & 5:26           & 46\%          & 75\%          & 0,95              \\
Jack Up Car           & 89               & 3               & 4:13           & 39\%          & 81\%          & 1,00              \\
Make Jello Shots      & 182              & 6               & 4:15           & 21\%          & 72\%          & 0,87              \\
Change Tire           & 99               & 11              & 4:52           & 27\%          & 62\%          & 0,97              \\
Make Lemonade         & 131              & 8               & 3:44           & 28\%          & 69\%          & 0,80              \\
Add Oil to Car        & 137              & 8               & 5:39           & 33\%          & 85\%          & 0,92              \\
Make Latte            & 157              & 6               & 3:52           & 43\%          & 71\%          & 0,89              \\
Build Floating Shelves         & 153              & 5               & 5:23           & 34\%          & 58\%          & 0,96              \\
Make Taco Salad       & 170              & 8               & 4:44           & 41\%          & 79\%          & 0,66              \\
Make French Toast     & 252              & 10              & 4:10           & 23\%          & 68\%          & 0,80              \\
Make Irish Coffee     & 185              & 5               & 3:13           & 13\%          & 74\%          & 0,77              \\
Make Strawberry Cake  & 86               & 9               & 5:36           & 25\%          & 63\%          & 0,82              \\
Make Pancakes         & 182              & 8               & 4:34           & 19\%          & 70\%          & 0,89              \\
Make Meringue         & 154              & 6               & 4:42           & 23\%          & 67\%          & 0,98              \\
Make Fish Curry       & 149              & 7               & 5:31           & 25\%          & 69\%          & 0,74              \\ \midrule
\textbf{Average}      & \textbf{153}     & \textbf{7}      & \textbf{4:57}  & \textbf{31\%} & \textbf{72\%} & \textbf{0,86}     \\ \bottomrule
\end{tabular}
}
\end{table*}

\section{Dataset}
\label{sec:dataset}
\subsection{Video collection}
As described in Section \ref{subsec:video_collection} of the main paper, given a task, we collect top $N$ videos from YouTube, by querying a title of the task.
The choice of $N$ relies on the following trade-off.
Training the model on a large number of videos for a given task may lead to better performance.
On the other hand, large $N$ results in many videos being unrelated to the queried task, which may hurt the performance of the model.
To investigate the influence of $N$ on the purity of the data, we have annotated videos from YouTube search output as relevant or irrelevant to a task for all primary tasks.
We define the average relevance as the ratio between the number of relevant videos and the total number of videos.
Figure \ref{fig:relevance} shows the average relevance for different values of $N$.
The relevance rapidly decreases with $N$, making the data unusable without manual cleaning.
For each related task we take top 30 videos from YouTube, which seems a reasonable compromise between the amount of data and the level of noise.

30 videos are clearly not enough to learn a task from scratch, as they feature only a few positive examples for each step in the task. Sharing knowledge across tasks as proposed in our paper is an essential mechanism to overcome this problem.

Since the videos are collected automatically for each task, they may be shared between tasks.
The primary tasks have little in common and do not share any videos. The related tasks, however, may be similar to each other and to the primary tasks (\eg \textit{Make Sourdough Pancakes} and \textit{Make Pancakes}).
We found that the primary and related tasks share about 2.6\% of videos.
However, we stress the fact that such duplicate videos are provided with different supervision (different ordered list of steps) for each task.
Our videos come from 3007 different YouTube channels, with $1.6$ videos per channel on average. $70\%$ of videos from primary tasks do not share channels with videos from related tasks. We, thus, conclude, that the difference between videos collected for primary and related tasks is sufficiently large,
hence transferring related task models to primary tasks is not trivial.

\subsection{Tasks and statistics}
Figure \ref{fig:tasks} shows all 83 primary and related tasks from our dataset.
In order to illustrate the sharing between tasks, we define the distance between two tasks based on the number of common step components and make a 2D projection via t-SNE \citesup{maaten08tsne}.
The tasks mostly share within the same domain, forming different groups, such as \textit{Car Maintenance}, \textit{Drinks} and \textit{Woodworks}.
\textit{Car Maintenance} and \textit{Home Woodwork} tasks mostly share components within the same category, with some spurious sharing with cooking tasks (\eg \textit{Pour} and \textit{Oil} components from \textit{Add Oil to Car} task).
This categories form two distinct clusters on the diagram.
The rest of the tasks form a continuous \textit{Cooking} cluster.

Table \ref{tab:datastat} provides some statistics for the primary tasks.
The order consistency, defined as in \citesup{alayrac:hal-01580630}, shows how well the order is respected in the videos. For example, if the order of steps in a video is $2\to 1\to 3$, the order consistency for this video would be equal to $\frac23$.
The amount of background is defined as an average number of frames, which are not assigned to any step, divided by a total number of frames. The high amount of background (72\% in average) motivates the use of methods, that aren't limited to a dense segmentation of a video and allow frames to remain unlabeled.
The average order consistency is high (86\%), thus justifying the use of hard ordering constraints. However, it varies across the tasks and has a relatively low value for some of the tasks (\eg \textit{Make Kimchi Rice} and \textit{Make Taco Salad}).
Similarly, the amount of missing steps is close to 50\% for \textit{Grill Steak} and \textit{Pickle Cucumber}, making these tasks especially challenging for our method.

\section{Experiments}
\label{sec:experiments}

\subsection{Comparison of evaluation metrics}
\label{subsec:metrics}
At test time we predict one temporal unit per step and assume a correct detection if it falls within a ground truth interval for the corresponding step. This is motivated by the fact that in weakly supervised context, when no information about exact temporal extents of steps is given during the training, prediction of step time intervals is an ill-posed problem. Indeed, even people do not always agree on the action boundaries. Predicting punctual steps, defined as the most consistent and distinguishable frames in the videos, allows to avoid this problem.
Although our model is trained for this punctual prediction, it may still be used to predict temporally extended steps, for example, by thresholding model's confidence of each step for each frame. Does this yield reasonable predictions? To answer this question we evaluate our model, using mAP metric. Table \ref{tab:map} contains results, averaged over the primary tasks and compared to baselines. Here recall stands for the same evaluation procedure, as described in the main paper (global non-maximal suppression + recall).
Note that both proposed ways of evaluation yield highly correlated results with recall roughly equal to $\text{mAP}\times 2$. The only exception is Richard'18 that under-performs in the case of recall. This may be caused by the fact that, unlike other methods in Table \ref{tab:map}, it is trained to predict step intervals, and not punctual steps.

\begin{table}[t]
\caption{Results of cross-task learning, evaluated with mAP and recall metrics and averaged over primary tasks. Standard deviation does not exceed 0.3\% for mAP and 1\% for recall.}
\label{tab:map}
\resizebox{0.48\textwidth}{!}{
\begin{tabular}{@{}lccccc@{}}
\toprule
Metric & Random & Richard'18 & Alayrac'16 & \begin{tabular}{c} Ours \\ (no sharing)\end{tabular} & \begin{tabular}{c} Ours \\ (with sharing)\end{tabular} \\
\midrule
Recall & 8.27 & 6.7 & 13.3 & 18.6 & 22.4 \\
mAP & 4.3 & 5.5 & 6.9 & 8.9 & 11.0 \\
\bottomrule
\end{tabular}
}
\end{table}

\subsection{Importance of temporal text constraints}
\label{subsec:constraints}
Temporal constraints, obtained from narration, provide a very noisy supervision.
In case of our primary tasks, the intersection over union between the ground truth of the steps and the corresponding temporal text constraints is only $7.9\%$. Moreover, $61\%$ of ground truth steps lie entirely outside of the constraint intervals.
This means that our method is unable to assign a step to a correct frame even with a perfect classifier, if it is forced to satisfy these constraints.
This is likely to be a disadvantage during training which tries to fit the step model to incorrect temporal intervals.
Could it be better to learn our model without temporal constraints?
We answer this question by training and evaluating our solution in the same setup as before, but without text constraints at the training time.
The resulting recall is $17\%$, compared to $22.4\%$, when training with text constraints.
This gain of $5.4\%$ shows the importance of text guidance during training even at the presence of the high level of noise.

\subsection{Additional qualitative results}
\label{subsec:results}
\begin{figure*}[t]
\begin{center}
   \includegraphics[width=\linewidth]{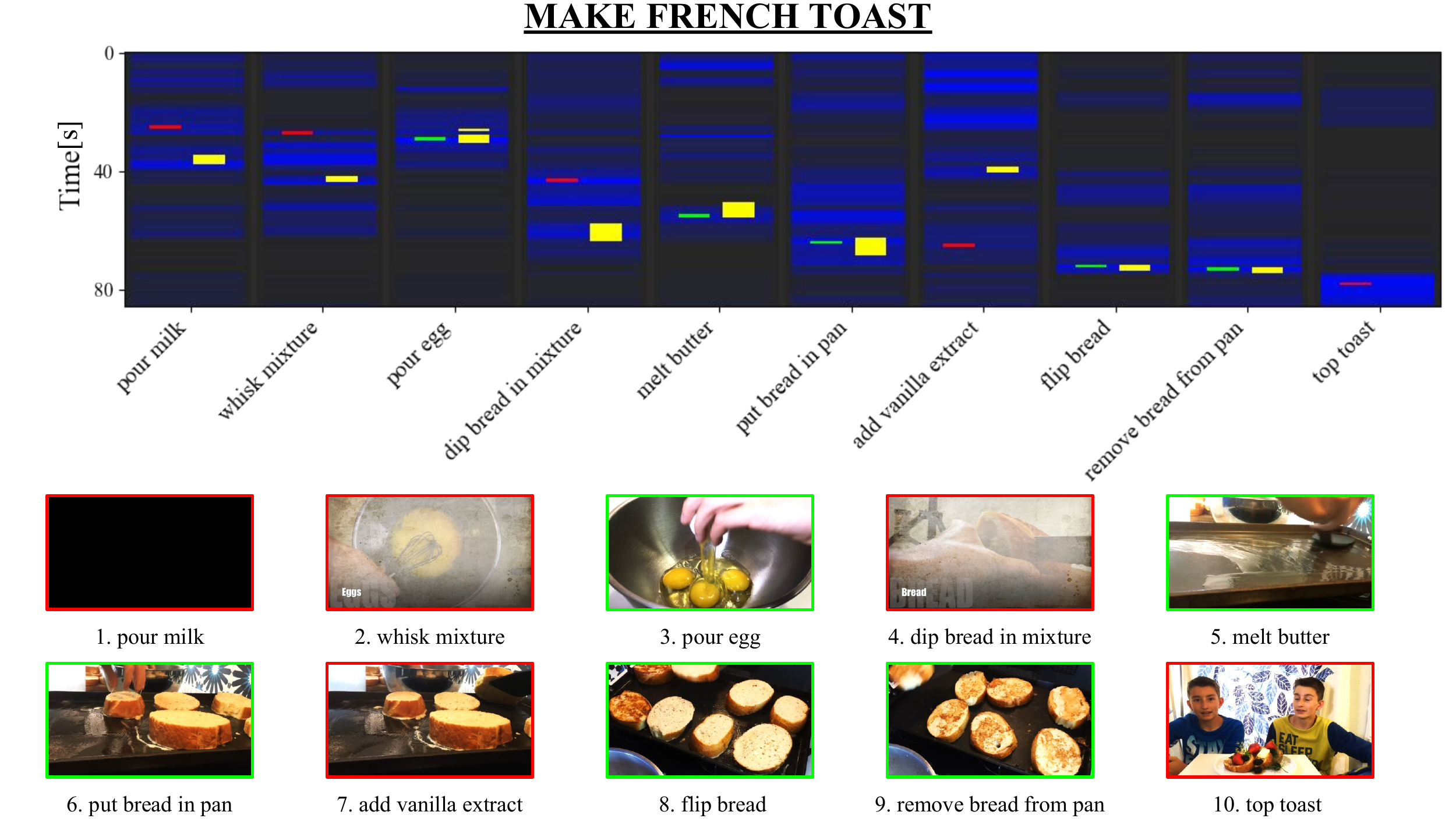}
   \caption{Example of obtained solution for \textit{Make French Toast} task. Outputs of the classifier are shown in \textcolor{blue}{blue}. Correctly localized steps are shown in \textcolor{green}{green}. False detections are shown in \textcolor{red}{red}. Ground truth intervals for the steps are shown in \textcolor{yellow}{yellow}. Failure cases include false localization due to the ordering constraints (\textit{Pour milk}, \textit{Whisk mixture} and \textit{Dip bread}) and due to a missing step (\textit{Top toast}).}
   \label{fig:assignment_toast}
\end{center}
\vspace{-0.2in}
\end{figure*}

\begin{figure*}[t]
\begin{center}
   \includegraphics[width=\linewidth]{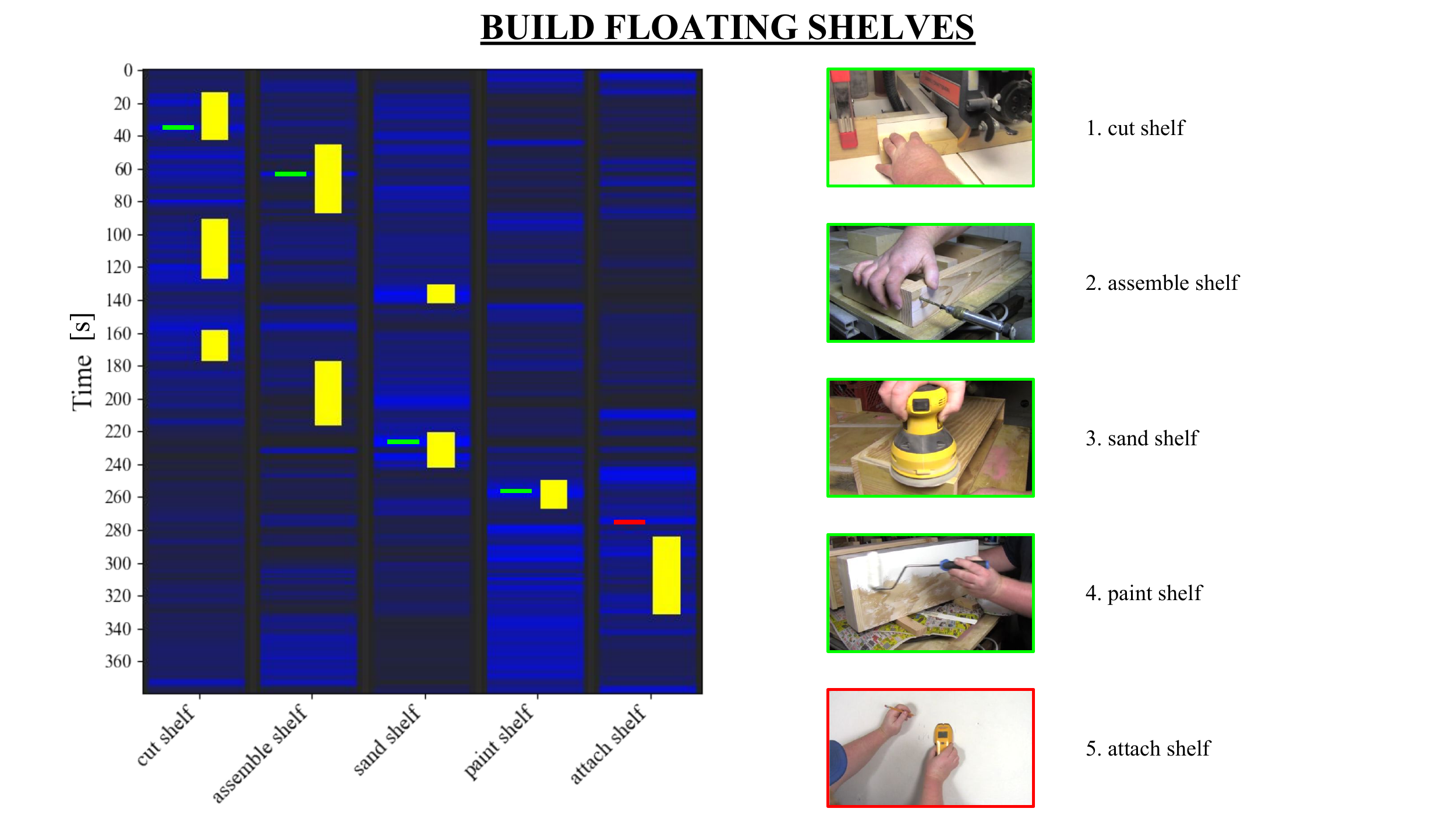}
   \caption{Example of obtained solution for \textit{Build Floating Shelves} task. Outputs of the classifier are shown in \textcolor{blue}{blue}. Correctly localized steps are shown in \textcolor{green}{green}. False detections are shown in \textcolor{red}{red}. Ground truth intervals for the steps are shown in \textcolor{yellow}{yellow}.}
   \label{fig:assignment_shelf}
\end{center}
\vspace{-0.2in}
\end{figure*}

\begin{figure*}[t]
\begin{center}
   \includegraphics[width=\linewidth]{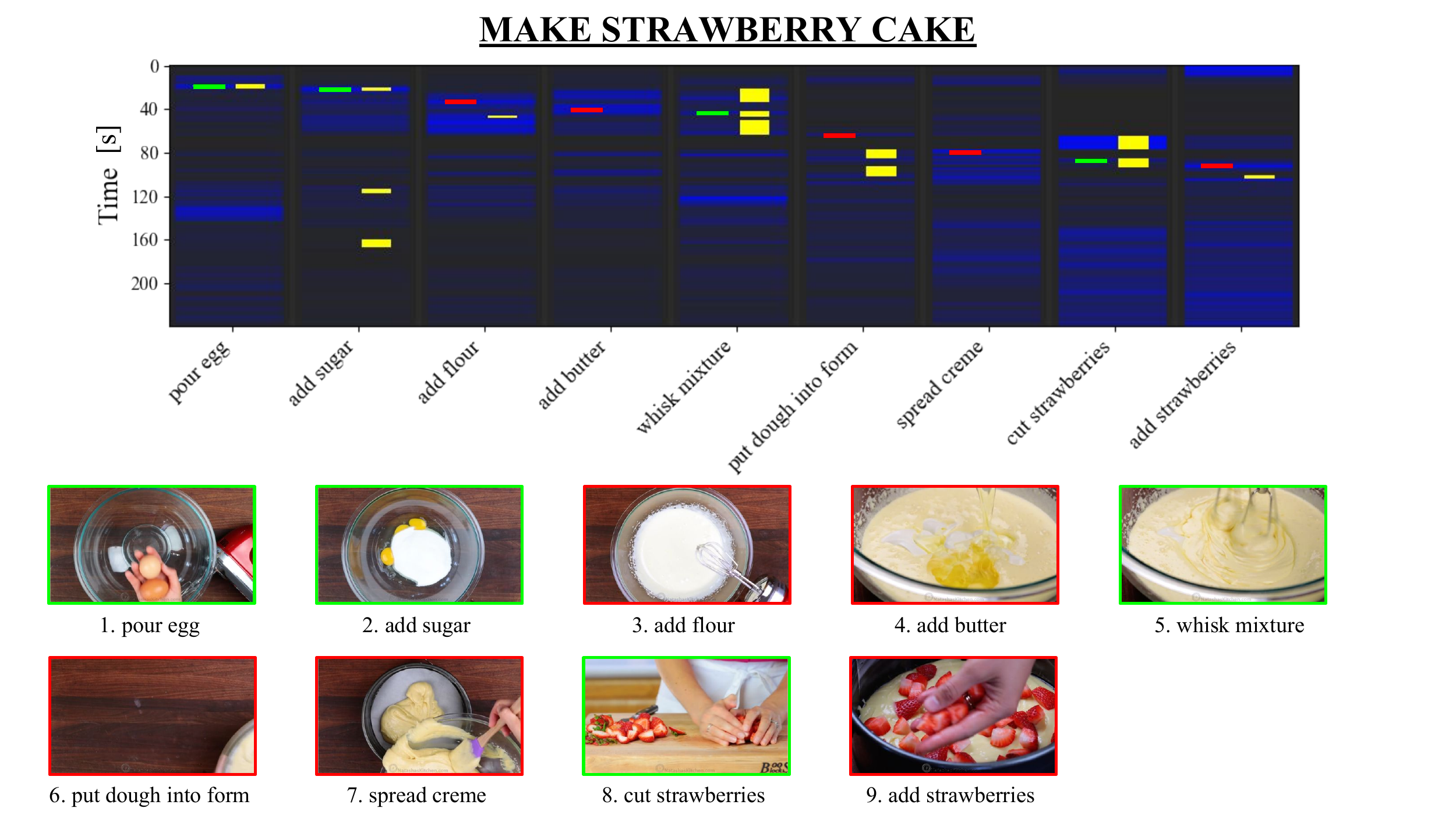}
   \caption{Example of obtained solution for \textit{Make Strawberry Cake} task. Outputs of the classifier are shown in \textcolor{blue}{blue}. Correctly localized steps are shown in \textcolor{green}{green}. False detections are shown in \textcolor{red}{red}. Ground truth intervals for the steps are shown in \textcolor{yellow}{yellow}. }
   \label{fig:assignment_cake}
\end{center}
\vspace{-0.2in}
\end{figure*}

\begin{figure*}[t]
\begin{center}
   \includegraphics[width=\linewidth]{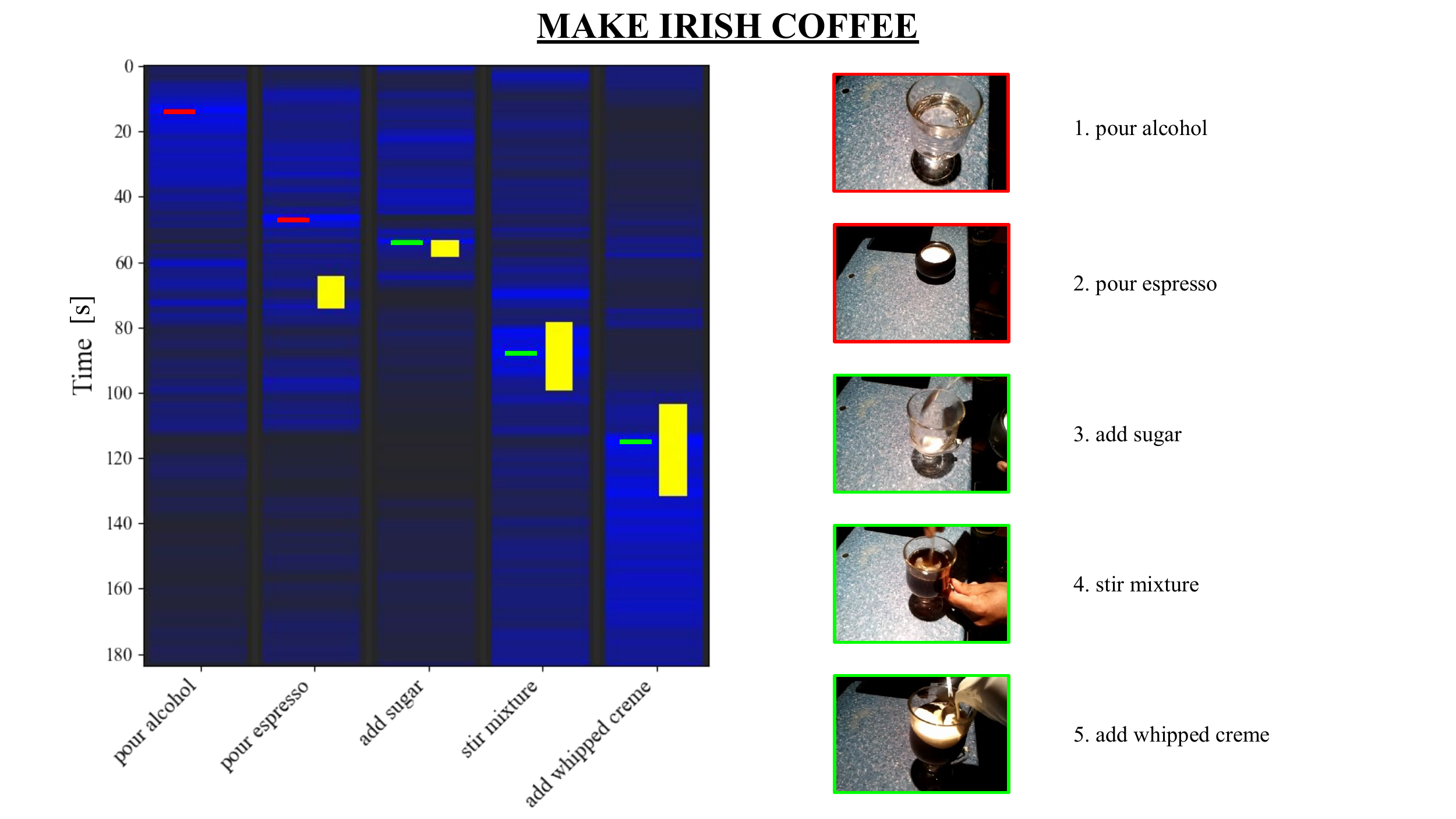}
   \caption{Example of obtained solution for \textit{Make Irish Coffee} task. Outputs of the classifier are shown in \textcolor{blue}{blue}. Correctly localized steps are shown in \textcolor{green}{green}. False detections are shown in \textcolor{red}{red}. Ground truth intervals for the steps are shown in \textcolor{yellow}{yellow}.}
   \label{fig:assignment_irish}
\end{center}
\vspace{-0.2in}
\end{figure*}

\begin{figure*}[t]
\begin{center}
   \includegraphics[width=\linewidth]{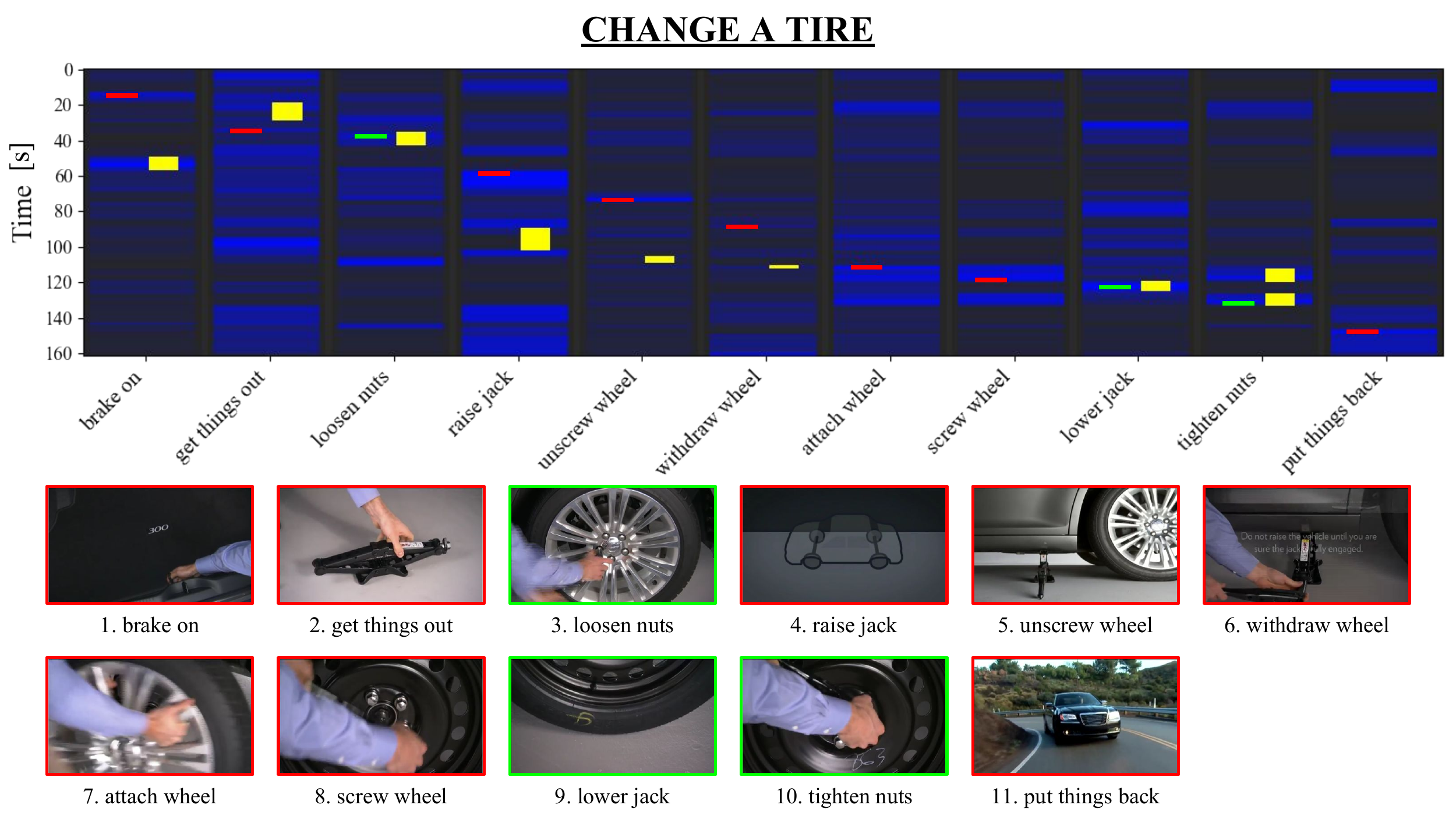}
   \caption{Example of obtained solution for \textit{Change a Tire} task. Outputs of the classifier are shown in \textcolor{blue}{blue}. Correctly localized steps are shown in \textcolor{green}{green}. False detections are shown in \textcolor{red}{red}. Ground truth intervals for the steps are shown in \textcolor{yellow}{yellow}.}
   \label{fig:assignment_tire}
\end{center}
\vspace{-0.2in}
\end{figure*}

\begin{figure*}[t]
\begin{center}
   \includegraphics[width=\linewidth]{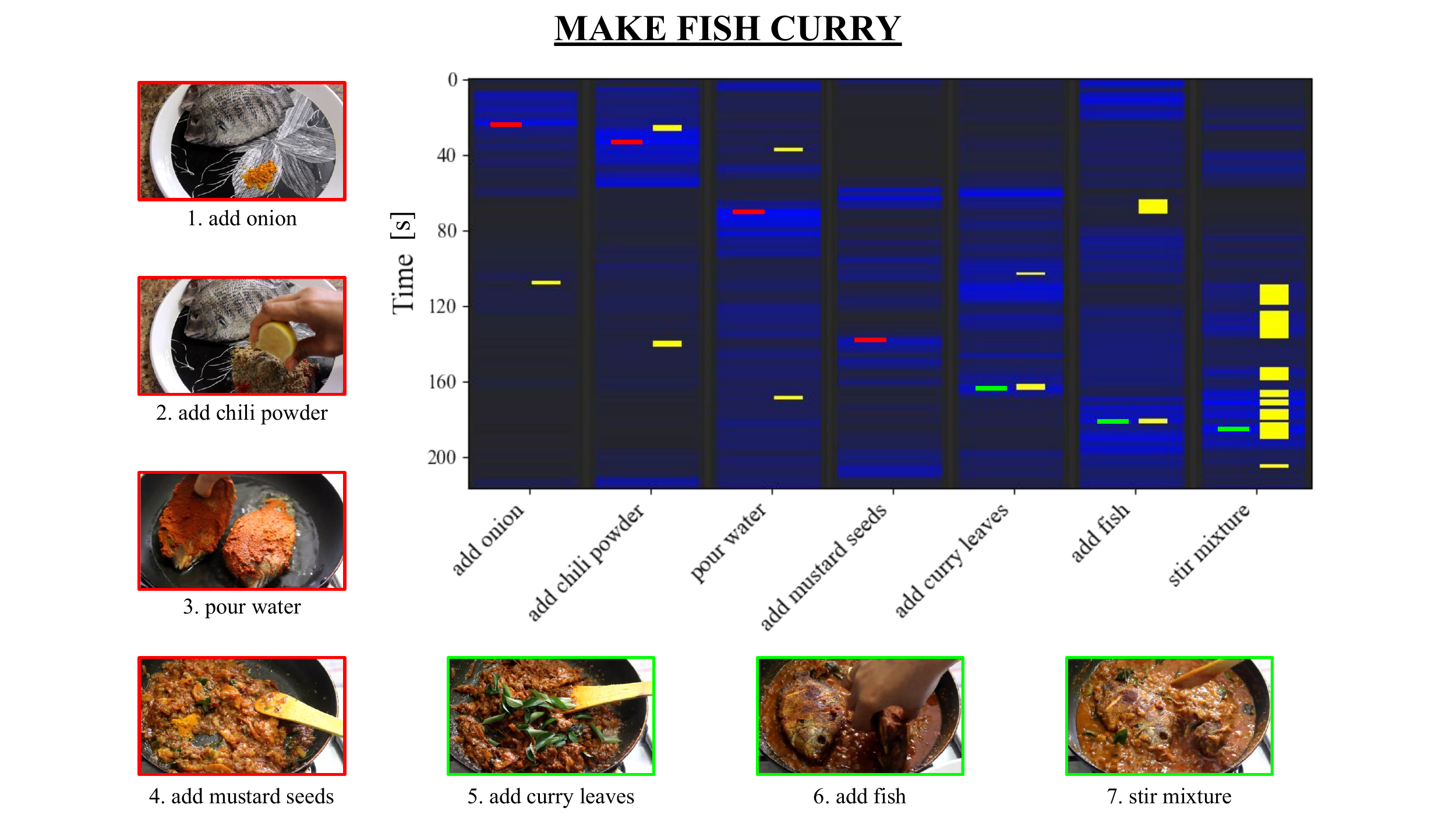}
   \caption{Example of obtained solution for \textit{Make Fish Curry} task. Outputs of the classifier are shown in \textcolor{blue}{blue}. Correctly localized steps are shown in \textcolor{green}{green}. False detections are shown in \textcolor{red}{red}. Ground truth intervals for the steps are shown in \textcolor{yellow}{yellow}.}
   \label{fig:assignment_fish}
\end{center}
\vspace{-0.2in}
\end{figure*}

In this section we show some additional qualitative examples to provide a better intuition into our data and the method.
Figures \ref{fig:assignment_toast}-\ref{fig:assignment_fish} shows the outputs of our model for videos for several tasks.
We show the outputs of classifiers for each step at each frame of the video, as well as the inferred solution and compare it with the ground truth.
Figure \ref{fig:assignment_toast} illustrates two kinds of error, caused by our assumptions.
First,the step \textit{Whisk Mixture} is localized in the area of low confidence for this step.
This is caused by the next step, \textit{Pouring Egg}, that precedes \textit{Whisking Mixture} in this particular video.
Second, false detection for \textit{Topping Toast} is due to the absence of this step in the video, while we assume that every step is present.
Note that although step \textit{Top Toast} doesn't appear in the video, the classifier puts high and well localized scores in the end of the video.
This is because \textit{Making French Toast}define positive as falling inside a ground truth interval for that step videos usually end with a demonstration of a final product on a plate.
The model captures this visual consistency between the videos and takes it for the final step.

\begin{figure*}[t]
\begin{center}
   \includegraphics[width=\linewidth]{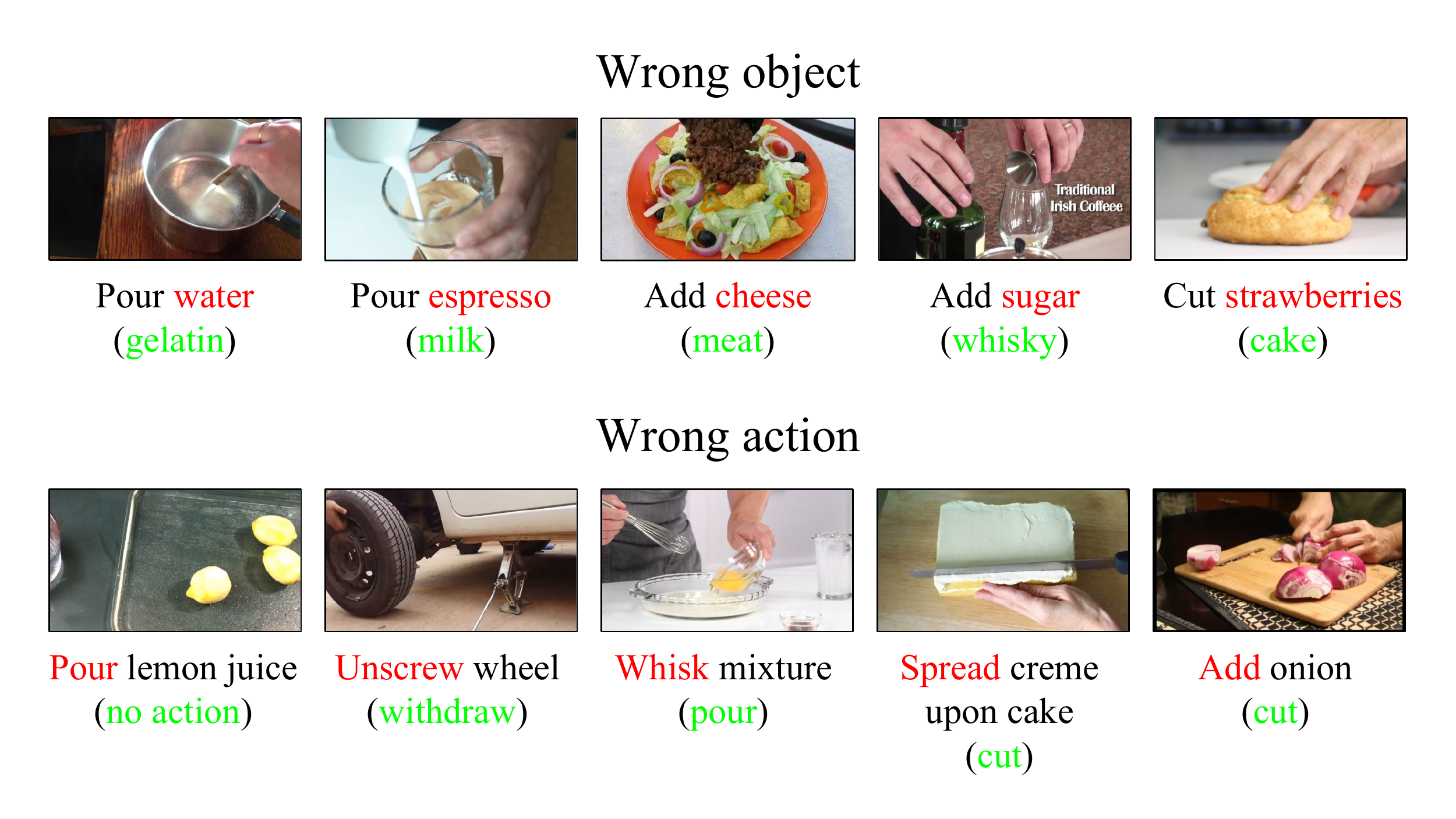}
   \caption{Erroneous predictions, involving wrong objects and actions. Correct object/action is in \textcolor{green}{green}. Our method is not capable of distinguishing particular kinds of objects, especially liquids and powders, due to the nature of the features. Examples for the wrong action components show that in many cases the method captures a static context in which object occurs, rather than performed action.}
   \label{fig:failures}
\end{center}
\vspace{-0.2in}
\end{figure*}

{\small
\bibliographystylesup{ieee}
\bibliographysup{biblio}
}

\end{document}